\pdfoutput=1 %
\documentclass[acmtog,nonacm,screen]{acmart}
\usepackage{paper}
\bqreviewfalse %
\graphicspath{{figures/arxiv/}{figures/}}
\makeatletter\let\bq@authorline\@typeset@author@line\def\@typeset@author@line{\raggedright\bq@authorline}\makeatother
\begin{document}
\title{Sharing Structure, Not Decisions: Batched Incremental Potential Contact for Robot Rollouts}
\author{Zhaofeng~Luo}
\affiliation{\institution{Carnegie Mellon University}\country{USA}}
\author{Xinyu~Lu}
\affiliation{\institution{Genesis AI}\country{USA}}
\author{Lalit~Jayanti}
\affiliation{\institution{Carnegie Mellon University}\country{USA}}
\author{Alexis~Duburcq}
\author{Tsun-Hsuan~Wang}
\author{Yi-Ling~Qiao}
\affiliation{\institution{Genesis AI}\country{USA}}
\author{Minchen~Li}
\affiliation{\institution{Carnegie Mellon University}\country{USA}}
\affiliation{\institution{Genesis AI}\country{USA}}
\authorsaddresses{}

\begin{abstract}
Evaluating, training, and optimizing robot policies for deformable manipulation consumes enormous numbers of independent, contact-rich trajectories. Incremental potential contact (IPC) makes these trajectories robust, but every step nests Newton iterations, Krylov solves, and continuous collision detection, and this work differs from trajectory to trajectory. We present \system, a GPU system that advances thousands of IPC simulations together under one principle: \emph{share structure, not decisions}. Environments with repeated topology share one read-only template of connectivity, sparsity patterns, collision hierarchies, and preconditioner maps, and store their private values side by side in memory; each environment owns its Newton iterations, line search, Krylov recurrences, collision bounds, and physical clock, and a finished environment rejoins the batch at the next Newton round. Under a strict convergence and intersection-free contract, \system sustains up to 36,000 environment steps per second on one H100 GPU, 4.9--10.0\x the throughput of the best multi-process configuration of a CUDA-graph implementation of libuipc and 7.3--8.2\x that of libuipc itself. The speedup combines fewer Newton iterations, because no environment waits for another's convergence test, with cheaper iterations, because all environments form one wide batch over a shared template; giving a conventional tiled solver per-environment convergence decisions alone closes only a small fraction of the gap. In our tests, batching perturbs trajectories and task outcomes no more than rerunning an environment alone does. The gains are largest for small deformable environments that progress at different rates, vanish when one environment fills the GPU, and carry over to policy evaluation, trajectory optimization, and reinforcement learning, which run under each task's iteration budgets instead of the strict contract.
\end{abstract}


\begin{teaserfigure}
\centering
\includegraphics[width=\textwidth]{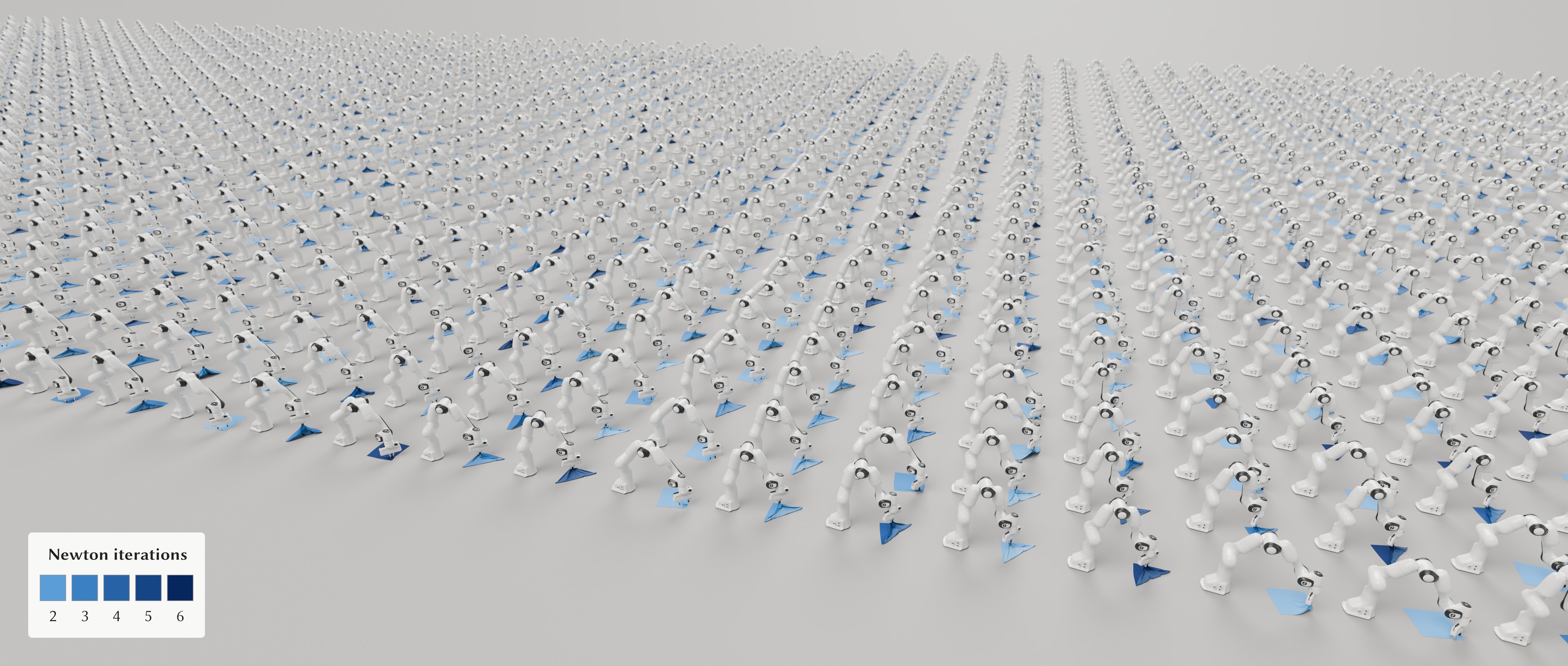}
\caption{\textbf{Same structure, different numerical work.} All 2,048 rollouts of one evaluation batch of a cloth-folding policy at the same simulated time, $t{=}4$\,s, rendered from the states \system computed. They share the robot, the $31{\times}31$ cloth mesh, and the solver, yet are in different contact states: each cloth is tinted by the number of Newton iterations its rollout needed in this step: two (885 rollouts), three (249), four (398), five (248), or six, the task's per-step budget (268; over the episode, 2.8\% of this evaluation batch's steps end at this budget without meeting the stopping test). \system shares the structure of such rollouts but leaves every numerical decision to each rollout, so that no rollout's decisions depend on another's. The policy was trained with PPO for the cloth-fold task of \cref{sec:applications}.}
\Description{A wide render of thousands of white robot arms standing in rows on a floor, each folding a cloth tinted in one of five shades of blue by its Newton iteration count, receding into the distance.}
\label{fig:teaser}
\end{teaserfigure}

\maketitle
\hypersetup{pdfsource={},pdfsubject={Computing methodologies: Physical simulation; Massively parallel algorithms; Computer systems organization: Single instruction, multiple data}, pdfkeywords={incremental potential contact, batched simulation, GPU, deformable manipulation, robot learning, policy evaluation}}
\pdfinfo{/Keywords (incremental potential contact, batched simulation, GPU, deformable manipulation, robot learning, policy evaluation)}
\pdfinfo{/Author (Zhaofeng Luo, Xinyu Lu, Lalit Jayanti, Alexis Duburcq, Tsun-Hsuan Wang, Yi-Ling Qiao, Minchen Li)}
\section{Introduction}
\label{sec:introduction}

Robot learning has turned physics simulation into a throughput problem. Estimating whether a policy folds a towel reliably takes thousands of episodes; a reinforcement-learning run consumes millions of them; a trajectory optimizer evaluates populations of candidate motions at every iteration. GPU simulators answer this demand by advancing thousands of independent environments at once~\citep{makoviychuk2021isaacgym,shacklett2023madrona,mujocowarp,newtonphysics}. Their solvers, however, were designed for rigid bodies and articulations. Deformable manipulation---grasping, folding, and wrapping cloth, pressing soft objects, routing cables---is where contact is most intricate and where simulation artifacts most easily corrupt what a policy learns or how it is scored.

Incremental potential contact (IPC) provides the robustness that these tasks require: its barrier formulation keeps every state free of intersections under large time steps, across volumetric solids, codimensional cloth, and stiff affine bodies~\citep{li2020ipc,li2021cipc,lan2022abd}. That guarantee has a price. Each time step minimizes a nonconvex incremental potential with several Newton iterations, each requiring a preconditioned conjugate-gradient (PCG) solve, a continuous collision detection (CCD) pass, and an energy-based line search. GPU implementations have made a single such solve fast~\citep{huang2024gipc,huang2025stiffgipc,wang2026agipc}, but a rollout workload needs \emph{many} of them, and they are not alike.

\begin{figure}[t]
\centering
\includegraphics[width=\columnwidth]{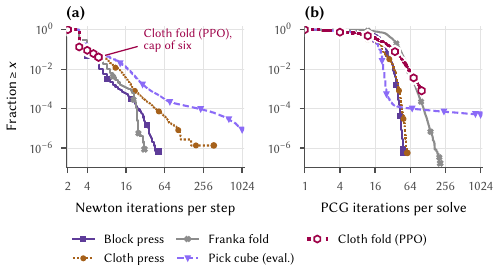}
\caption{\textbf{Work per step is long-tailed.} Complementary cumulative distributions of (a) Newton iterations per time step and (b) PCG iterations per linear solve, over all environments and steps of the three benchmarks and two applications. Newton counts reach 51 on the block press, 371 on the cloth press, and the budget of 1,024 in the pick-cube evaluation; the cloth-fold policy task caps Newton at six and PCG at 100 iterations, and its curve is one evaluation batch of a timed training run's final policy (0.08\% of its solves reach the PCG cap; 0.05\% over the timed training runs, Appendix~\ref{app:applications}). A batch of thousands almost always contains an environment in the tail.}
\Description{Two log-scale complementary cumulative distribution plots. Newton iteration counts decay slowly to beyond 30 iterations; PCG iteration counts spread over two orders of magnitude.}
\label{fig:iterations}
\end{figure}

Consider the cloth-folding rollouts in \cref{fig:teaser}. They share the robot and the cloth mesh, so their topology, sparsity patterns, and collision hierarchies are identical. Their numerical work is not: at the same instant, the 2,048 rollouts of this batch need between two and six Newton iterations per step---six is the task's budget---and each rollout's costly contact phases fall at different times. Across the five workloads we profiled, per-step Newton iteration counts have long tails, and PCG iteration counts vary by two orders of magnitude across solves (\cref{fig:iterations}). Repeated rollouts are regular in structure and irregular in progress.

\begin{figure}[t]
\centering
\includegraphics[width=\columnwidth]{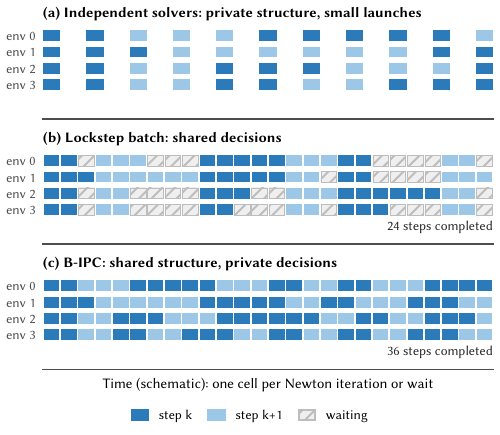}
\caption{\textbf{Share structure, not decisions.} Three ways to execute a batch whose Newton iteration counts differ per step, over the same time (shades mark consecutive physical steps). (a) Independent solvers keep decisions private but duplicate structure and issue small launches, drawn as gaps of arbitrary width, so (a) gives no step count. (b) A lockstep batch issues large launches but every environment waits for the slowest one. (c) \system pools stages, keeps decisions private, and refills finished environments at the next Newton round.}
\Description{Three timelines of four environments. In the first, blocks are separated by gaps. In the second, many slots are hatched as waiting. In the third, all slots are filled.}
\label{fig:design-space}
\end{figure}

The two obvious ways to run them both waste this property (\cref{fig:design-space}). Running an independent solver per environment respects each rollout's progress, but duplicates the structural data, issues many small kernels that leave the GPU underoccupied, and reads every environment's indices separately. Merging all environments into one large scene produces large kernels---but binds the environments to shared convergence and line-search decisions, so that the slowest rollout dictates the iteration count of all, and a single batch-wide residual test can either oversolve easy environments or accept unconverged hard ones.

\paragraph{Our approach.} Our system, \system (batched IPC), resolves this tension with one organizing principle: \emph{share structure, not decisions}. Everything that rollouts of a repeated scene have in common---mesh connectivity, block-sparse patterns, collision-tree links, preconditioner hierarchies---is stored once as a read-only template. Everything that makes a rollout itself---positions, contacts, Hessian values, preconditioner factors, Newton and line-search state, PCG scalars, collision bounds, and the physical clock---is private to its environment. Execution is pooled at the granularity of solver \emph{stages}: one assembly, one PCG, and one CCD stage serve all environments that currently need them, each with its own values and its own stopping decision. An environment that completes a time step starts its next step at the following Newton round instead of waiting for the slowest environment.

Three design decisions make this principle efficient on a GPU (\cref{sec:system}). First, private arrays place the environment index innermost, so that adjacent threads process the same structural item in different environments and share one index fetch---the repeated topology becomes memory coalescing. Second, a hybrid operator keeps the fixed-pattern part of each Hessian in a shared block-sparse layout and applies the changing contact terms as private low-rank factors, avoiding per-step reassembly of a contact-dependent sparsity pattern. Third, the nonlinear control flow is segmented by environment and driven by compacted lists of active work, so that pooled execution never couples the decisions of two rollouts.

\paragraph{Results.} Under a strict contract---every counted step converged and intersection-free---\system delivers 4.9--10.0\x the throughput of \cgq's best multi-process configuration on one H100 and 7.3--8.2\x that of the public libuipc library~\citep{libuipc} (\cref{sec:eval-throughput}). The gain has two sources: per-environment decisions avoid unnecessary Newton and PCG work, while one wide batch over a shared template makes each remaining iteration cheaper. Giving \cgq the same per-environment convergence decisions closes only a small fraction of the gap, showing that local stopping alone does not explain the result (\cref{sec:eval-throughput,sec:eval-attribution}). We also compare complete systems with \taccel~\citep{li2025taccel} and NVIDIA Newton to place the IPC throughput in context (\cref{sec:eval-cross}). Across policy evaluation, trajectory optimization, and reinforcement learning, the speedup translates into earlier estimates, solutions, and training progress (\cref{sec:applications}). The gains are largest for small deformable environments that progress at different rates, shrink on self-contact-dominated workloads, and vanish once a single environment fills the GPU (\cref{sec:eval-generality}).

\paragraph{Contributions.} In summary, this paper makes the following contributions:
\begin{itemize}[leftmargin=*,topsep=2pt,itemsep=1pt]
\emergencystretch=1em %
\item \emph{Decision-local batching for nested IPC solves}: we pool each solver stage across environments while keeping Newton and PCG convergence, line search, collision bounds, barrier updates, and time advancement local to each environment. We give checkable ownership and reduction rules under which batching preserves the single-environment computation in exact arithmetic, with floating-point differences limited to rounding (\cref{prop:invariance,sec:determinism}).
\item The design of \system: we store repeated scene structure once, place the environment index innermost in private arrays, split the Hessian into shared-pattern and private low-rank contact terms, segment PCG and collision processing by environment, and refill completed environments at Newton-round granularity (\cref{sec:system}).
\item An evaluation against four GPU simulators that separates numerical work from execution cost, attributes the speedup to its components, maps the regime in which batching wins, and measures end-to-end effects on policy evaluation, trajectory optimization, and reinforcement learning (\cref{sec:evaluation,sec:applications}).
\end{itemize}
We will release \system together with its kernel-language compiler, the benchmark scenes, and all measurement scripts.

\section{Related Work}
\label{sec:related}

\paragraph{Incremental potential contact.}
IPC formulates implicit time stepping with frictional contact as the minimization of an incremental potential augmented by a log-barrier on geometric distances, and filters every Newton update with continuous collision detection so that iterates stay intersection-free~\citep{li2020ipc}. Codimensional IPC extends the formulation to shells and rods with thickness offsets and additive CCD~\citep{li2021cipc}; rigid IPC and affine body dynamics treat stiff and rigid objects with intersection-free piecewise-linear trajectories~\citep{ferguson2021rigidipc,lan2022abd}. Barrier variants change the stiffness schedule or the barrier itself~\citep{ando2024cubic,guo2024bal}; penetration-free elastodynamics can also be enforced without barriers through constrained optimization~\citep{zheng2026barrierfree}. \system does not alter these formulations: it executes the potentials, barriers, friction, and stopping rules of its reference implementation, and changes only how many independent instances of them are organized on a GPU.

\paragraph{Fast GPU solvers for one contact problem.}
A large body of work accelerates the solve inside a single simulation. GIPC derives analytic barrier eigensystems for a fully GPU-resident IPC pipeline~\citep{huang2024gipc}; StiffGIPC adds a connectivity-enhanced multilevel additive Schwarz (MAS) preconditioner~\citep{wu2022mas} and affine--deformable coupling~\citep{huang2025stiffgipc}; the public libuipc library packages this line~\citep{libuipc}; AGIPC coarsens the linear system adaptively during the solve~\citep{wang2026agipc}; YASPS generates block-sparse IPC kernels from symbolic relational specifications~\citep{tang2026yasps}. Barrier-augmented Lagrangian methods and Jacobi- or Gauss--Seidel-style descent trade Newton's global solve for parallel local updates~\citep{guo2024bal,lan2023stencil,lan2025jgs2}, and vertex block descent (VBD) and its augmented and offset-geometry variants reach real-time rates with per-vertex updates and conservative displacement bounds~\citep{chen2024vbd,giles2025avbd,chen2025ogc}. These methods make one solve cheaper. \system is orthogonal: it exploits parallelism \emph{across} solves that share structure but not progress, and reuses the reference MAS preconditioner unchanged. A faster per-environment solver would compose with it; we compare against VBD as an alternative way to buy throughput (\cref{sec:eval-cross}).

\paragraph{Many-world simulation for robot learning.}
Isaac Gym moved rigid-body simulation and policy inference onto the GPU to remove host round trips~\citep{makoviychuk2021isaacgym}; Brax, MuJoCo Warp (a batched GPU port of MuJoCo~\citep{todorov2012mujoco}), ManiSkill3, Isaac Lab, Genesis, and Newton provide batched rigid and articulated dynamics for learning~\citep{freeman2021brax,mujocowarp,tao2025maniskill3,mittal2023orbit,genesis2024,newtonphysics}. Deformable objects in these frameworks are typically simulated with position-based or particle methods~\citep{macklin2014flex,lin2020softgym}, with the material point method~\citep{huang2021plasticinelab,gu2023maniskill2}, often written in Warp~\citep{macklin2022warp}, or with finite-element soft bodies and penalty contact, as in Isaac Lab, the successor of Orbit~\citep{mittal2023orbit}, all of which trade the guarantees of IPC for speed. Batched rigid-body solvers typically bound the iterations of every step; when a loop with a per-world tolerance is vectorized---for example as a vectorized while-loop in JAX~\citep{jax2018github}, which masks finished worlds---it runs until the slowest world converges, the lockstep behavior we analyze in \cref{sec:scheduling}. Madrona showed that a data-oriented engine organized around thousands of small worlds outperforms one that runs them as separate instances~\citep{shacklett2023madrona}. A GPU soft-robot simulator coupled to Newton's rigid-body solver likewise stores reduced bases, cubature rules, and collision proxies once for all environments, and advances them in lockstep with one linearly implicit solve and penalty contact per step~\citep{trusty2026gpusoftrobot}. \system brings this many-world perspective to implicit barrier contact, whose nested Newton--Krylov--CCD iterations make per-world work data-dependent and whose convergence decisions must not be shared.

\paragraph{Batched IPC\@.}
SAPIEN-IPC integrated IPC into a robot simulator~\citep{yuan2024sapienipc}, and IPC has modeled grasping and soft--rigid manipulation~\citep{kim2022ipcgraspsim,du2024softrigid}. \taccel combines IPC and affine bodies in a Warp implementation and reports thousands of parallel tactile-sensing environments~\citep{li2025taccel}; its conjugate-gradient solver already reduces dot products and tracks convergence per environment, and \cgq's partitioned PCG already ends the solve of each connected component of the Hessian, and thus of each environment of a tiled scene, on its own. We therefore claim neither the first batched IPC system nor the first per-environment Krylov solver.

\paragraph{Batched numerics and continuous batching.}
Several ingredients of \system have analogs outside simulation. Interleaved layouts put the batch index innermost for batched dense linear algebra~\citep{kim2017compact}; batched iterative solvers let small systems converge independently~\citep{aggarwal2021batched,aggarwal2022batched}; persistent and wavefront GPU methods refill irregular work as tasks finish~\citep{gupta2012persistent,aila2009traversal,laine2013megakernels}; batched sparse direct solvers reuse symbolic structure across matrices with a common pattern~\citep{nvidia2026cudss}; and continuous batching, batch-invariant kernels, and asynchronous environment engines address related scheduling or independence questions in language-model serving and reinforcement learning~\citep{yu2022orca,kwon2023vllm,he2025nondeterminism,petrenko2020samplefactory,weng2022envpool}. Our setting combines these concerns inside a nested nonlinear--linear--geometric solve: Newton, PCG, line search, collision handling, and time advancement all make data-dependent decisions. \system therefore couples shared templates and pooled GPU stages with an explicit per-environment correctness contract and Newton-round refill, and we attribute these choices end to end in \cref{sec:eval-attribution}.

\section{Batched Rollouts of Incremental Potential Contact}
\label{sec:contract}

We first state the computation \system performs and the contract it must honor, and then examine the space of designs that satisfy it.

\subsection{One environment: an IPC time step}
An \emph{environment} is one trial of a task: a robot, the objects it manipulates, and the static scene, all physically coupled. For environment $e\in\{1,\ldots,B\}$ in a batch of size $B$, let $\mathbf{x}_e\in\mathbb{R}^{n}$ collect its generalized coordinates---world positions of deformable vertices and the twelve affine coordinates of each stiff body~\citep{lan2022abd}. Robot links are affine bodies; joints are stiff quadratic potentials, and drives add an explicit force and a quadratic damping potential. With time step $h$, inertial prediction $\widetilde{\mathbf{x}}_e$, and mass matrix $\mathbf{M}_e$, backward Euler finds the next state by minimizing the incremental potential~\citep{li2020ipc}
\begin{equation}
  \Phi_e(\mathbf{x})=\tfrac{1}{2}\|\mathbf{x}-\widetilde{\mathbf{x}}_e\|_{\mathbf{M}_e}^{2}
  +h^2\bigl(\Psi(\mathbf{x};\boldsymbol{\theta}_e)+\kappa_e\,C_e(\mathbf{x})+F_e(\mathbf{x})\bigr),
  \label{eq:potential}
\end{equation}
where $\Psi$ contains elastic, joint, and drive potentials whose parameters $\boldsymbol{\theta}_e$ (controls, randomized materials) vary across environments, $C_e$ is the barrier potential summed over the environment's current contact pairs within activation distance $\hat d$, $\kappa_e$ is the environment's adaptive barrier stiffness, and $F_e$ is the friction potential, lagged once per time step from the previous step's normal forces. Each Newton iteration assembles the gradient $\mathbf{g}_e$ and a projected positive-definite Hessian $\mathbf{H}_e$, solves $\mathbf{H}_e\mathbf{d}_e=-\mathbf{g}_e$ approximately with MAS-preconditioned CG~\citep{wu2022mas} (PCG, our only Krylov solver), bounds the step length by CCD so that no contact pair crosses, and backtracks until $\Phi_e$ does not increase. The step terminates when
\begin{equation}
  \|\Delta\mathbf{x}_e\|_{\infty}<h\,\varepsilon_v,
  \label{eq:stopping}
\end{equation}
where $\Delta\mathbf{x}_e$ collects the world-space vertex updates that $\mathbf{d}_e$ induces before CCD scaling, i.e., when no vertex update implies a velocity change above the tolerance $\varepsilon_v$. \Cref{tab:parameters} lists the solver parameters; time steps are per scene (\cref{tab:benchmarks}).

\begin{table}[t]
\caption{Solver parameters, identical for \system and \cgq. Budgets that are exhausted make a step fail under C2. The Franka fold uses task-specific values, $\hat d{=}1$\,mm, $\kappa{=}10^4$\,N/m, $\mu{=}1$, $\varepsilon_v{=}0.05$\,m/s, and a PCG tolerance of $10^{-5}$ on both systems.}
\label{tab:parameters}
\centering\footnotesize
\setlength{\tabcolsep}{3pt}
\begin{tabularx}{\columnwidth}{@{}l l >{\raggedright\arraybackslash}X@{}}
\toprule
Parameter & Symbol & Value \\
\midrule
Barrier activation distance & $\hat d$ & 2\,mm \\
Barrier stiffness; growth bound & $\kappa_e$ & $10^5$\,N/m, adaptive per env.; $128{\times}$ \\
Friction; smoothing velocity & $\mu$; $\varepsilon_f$ & 0.5; $10^{-2}$\,m/s \\
Newton tolerance; budget & $\varepsilon_v$ & $10^{-2}$\,m/s; 1,024 iterations \\
PCG relative residual; budget & & $10^{-4}$; 1,024 iterations \\
Line-search trials; CCD gap kept & & ${\le}12$; $\eta{=}0.5$ of each pair's gap \\
Contact capacity & $c$ & preset per task; at least doubles on overflow \\
Precision & & FP64 state, operator, PCG; FP32 MAS \\
\bottomrule
\end{tabularx}
\end{table}

\subsection{Many environments: the batch contract}
\label{sec:batch-contract}
A batch of rollouts is $B$ such minimizations per time step. Because the environments are physically independent, the batched Newton system is block diagonal, $\operatorname{diag}(\mathbf{H}_1,\ldots,\mathbf{H}_B)\,[\mathbf{d}_e]_e=-[\mathbf{g}_e]_e$, but block-diagonal \emph{algebra} does not imply independent \emph{decisions}. A PCG dot product taken over the whole batch, a residual test on the stacked vector, a common line-search step, a shared barrier-stiffness update, a shared Newton-termination test, or a contact list or PCG reduction that mixes environments silently couples the environments: one rollout's trajectory would then depend on which others share its batch. For policy evaluation and learning, such coupling is unacceptable---success statistics would depend on batch composition, and a reset of one environment would perturb its neighbors. We therefore require \system to satisfy the following contract.

\begin{description}[leftmargin=0pt,labelindent=0pt,itemsep=2pt,topsep=3pt]
\item[C1 Batch invariance.] The trajectory of environment $e$ is determined by its own initial state and inputs: in exact arithmetic it does not depend on the batch size $B$ or on the states, inputs, resets, or deactivation of other environment slots, provided that no step exhausts the capacity-retry budget or reaches the index bound (\cref{sec:collision,sec:determinism}). In floating point, batching may change only rounding; because decisions are discontinuous, rounding differences can grow, so we require, and test in \cref{sec:eval-quality}, that batching perturbs a trajectory and its task outcome no more than rerunning the environment alone does.
\item[C2 Strict completion.] A step counts as completed only if it satisfies \cref{eq:stopping} within the budgets of \cref{tab:parameters}, its state is intersection-free (every contact pair at positive distance), and no capacity was exceeded. A step that fails is reported as failed and never counted as completed; whether the episode then continues from the failed state, as in our measurements, or is reset is left to the application.
\item[C3 Reference equivalence.] Within an environment, we use the same potentials, barrier and friction parameters, preconditioner, tolerances, and floating-point precision as the single-environment reference; batching may change execution and rounding, but not the numerical model or stopping rules.
\end{description}
The performance objective is the rate of strictly completed environment steps, in env-steps/s, at each system's best batch size $B^\star$. A \emph{window} is the sequence of steps an environment must complete before the host next intervenes---an episode, a control chunk, or a fixed horizon. \Cref{sec:eval-quality} verifies C1--C3.

\subsection{The design space}
\label{sec:design-space}

Per-step work is not only variable but long-tailed: Newton iteration counts decay slowly, so the maximum over a batch grows with its size (\cref{fig:iterations}). \Cref{fig:design-space} contrasts the two natural ways to execute such a batch. \emph{Independent solvers}---one solver instance, stream, or process per environment---satisfy C1 by construction, but a manipulation scene has only $10^3$--$10^4$ degrees of freedom: its kernels occupy a small fraction of a GPU, and each instance re-reads its own copy of indices, sparsity patterns, and collision trees. \emph{Lockstep batching} issues large kernels and shares structure, but either couples decisions (violating C1) or makes every environment wait for the slowest one in each step---and with long-tailed iteration counts, the slowest of thousands of environments is almost always deep in the tail. \system aims for both. We pool each solver stage across active environments while keeping every numerical decision local to its environment; we call this rule \emph{decision-local batching}. When an environment finishes a time step, we let it start the next one at the following Newton round instead of waiting for a batch-wide step boundary; we call this \emph{refill}. \Cref{sec:system} shows how we realize both ideas on the GPU.

\section{B-IPC: Shared Structure, Private Decisions}
\label{sec:system}

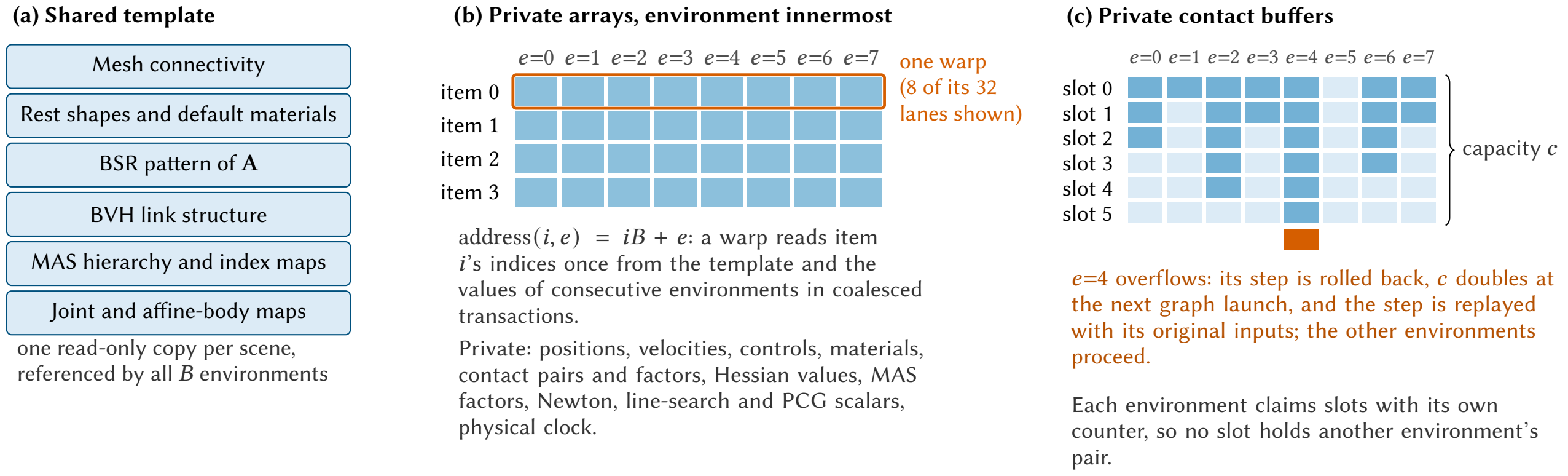
\begin{figure*}[t]
\centering
\begin{tikzpicture}[
  font=\sffamily\footnotesize, >={Stealth[length=4pt]},
  tbox/.style={draw=bqblue!70!black, fill=bqpale, rounded corners=1.5pt, minimum width=3.7cm,
               minimum height=0.47cm, align=center, inner sep=1.5pt},
  stage/.style={draw=bqgray, fill=white, rounded corners=1.5pt, minimum height=0.5cm,
                align=left, inner sep=2.5pt, text width=4.45cm},
  ptitle/.style={font=\sffamily\footnotesize\bfseries, anchor=west},
  note/.style={font=\sffamily\footnotesize, text=black!80, align=left,
               execute at begin node={\hyphenpenalty=10000\exhyphenpenalty=10000}},
]
\node[ptitle] at (0,2.6) {(a) Shared template};
\foreach \t [count=\i] in {Mesh connectivity, Rest shapes and default materials, BSR pattern of $\mathbf{A}$,
                           BVH link structure, MAS hierarchy and index maps, Joint and affine-body maps} {
  \node[tbox] at (1.9, 2.6-0.53*\i) {\t};
}
\node[note, anchor=north west] at (0.05,-0.75) {one read-only copy per scene,\\referenced by all $B$ environments};
\begin{scope}[xshift=4.75cm]
\node[ptitle] at (0,2.6) {(b) Private arrays, environment innermost};
\def\cw{0.5}\def\ch{0.36}
\foreach \e in {0,...,7} {
  \node[font=\sffamily\footnotesize, text=black!70] at (0.75+\e*\cw+\cw/2, 2.17) {$e{=}\e$};
}
\foreach \i in {0,...,3} {
  \node[font=\sffamily\footnotesize, anchor=east] at (0.72, 1.98-\i*\ch-\ch/2) {item $\i$};
  \foreach \e in {0,...,7} {
    \fill[bqblue!45] (0.75+\e*\cw+0.025, 1.98-\i*\ch-\ch+0.025) rectangle ++(\cw-0.05, \ch-0.05);
  }
}
\draw[bqorange, line width=0.9pt, rounded corners=1pt] (0.75, 1.98-\ch) rectangle ++(8*\cw, \ch);
\node[note, text=bqorange, anchor=west] at (0.75+8*\cw+0.05, 1.98-\ch/2) {one warp\\(8 of its 32\\lanes shown)};
\node[note, anchor=north west, text width=5.0cm] at (0.05,0.45) {$\mathrm{address}(i,e) = iB + e$: a warp reads item $i$'s indices once
  from the template and the values of consecutive environments in coalesced transactions.\\[2pt]
  Private: positions, velocities, controls, materials, contact pairs and factors, Hessian values, MAS factors,
  Newton, line-search and PCG scalars, physical clock.};
\end{scope}
\begin{scope}[xshift=11.35cm]
\node[ptitle] at (0,2.6) {(c) Private contact buffers};
\def\cw{0.42}\def\ch{0.27}
\foreach \e/\n in {0/3,1/1,2/5,3/2,4/6,5/0,6/4,7/2} {
  \node[font=\sffamily\scriptsize, text=black!70] at (0.75+\e*\cw+\cw/2, 2.17) {$e{=}\e$};
  \foreach \k in {0,...,5} {
    \ifnum\k<\n \def\fillc{bqblue!55}\else\def\fillc{bqpale}\fi
    \fill[\fillc] (0.75+\e*\cw+0.025, 1.98-\k*\ch-\ch+0.025) rectangle ++(\cw-0.05, \ch-0.05);
  }
}
\foreach \k in {0,...,5} { \node[font=\sffamily\footnotesize, anchor=east] at (0.72, 1.98-\k*\ch-\ch/2) {slot \k}; }
\draw[decorate, decoration={brace, amplitude=2.5pt}] (0.75+8*\cw+0.08, 1.98) -- node[right=2pt, note] {capacity $c$} (0.75+8*\cw+0.08, 1.98-6*\ch);
\fill[bqorange] (0.75+4*\cw+0.025, 1.98-6*\ch-0.04) rectangle ++(\cw-0.05, -0.22);
\node[note, text=bqorange!85!black, anchor=north west, text width=5.2cm] (ov) at (0.05,0.0) {$e{=}4$ overflows: its step is rolled back, $c$ doubles at the next graph launch, and the step is replayed with its original inputs; the other environments proceed.};
\node[note, anchor=north west, text width=5.2cm] at ([yshift=-2pt]ov.south west) {Each environment claims slots with its own counter, so no slot holds another environment's pair.};
\end{scope}
\end{tikzpicture}
\caption{\textbf{Share common structure; interleave private state.} (a) Quantities that rollouts of a scene have in common are stored once. (b) Fixed-size private arrays place the environment index innermost, so that a warp processes one structural item across consecutive environments with a single index fetch. (c) Contact buffers have one capacity $c$ for all environments, also environment-innermost; an environment that overflows is rolled back and replayed after $c$ grows, while the others proceed.}
\Description{Three panels: a list of shared template data structures, a grid of memory cells with items as rows and environments as columns with one row highlighted as a warp, and a grid of contact slots per environment with one environment overflowing.}
\label{fig:overview}
\end{figure*}

\system separates two concerns that existing designs conflate: \emph{who owns} a quantity and \emph{who executes} the work on it. Ownership follows the physics: a quantity that is identical across rollouts of a scene is owned by the scene's template; a quantity that can differ is owned by an environment (\cref{tab:ownership}). Execution follows the hardware: every solver stage runs once per Newton round over all environments that need it. \Cref{fig:overview} shows the resulting data organization and \cref{alg:round} the execution of a window.

\begin{table}[b]
\caption{Ownership rules. A quantity owned by an environment may live in shared buffers and be processed by shared kernels, but its value and every decision derived from it depend on that environment alone.}
\label{tab:ownership}
\centering\footnotesize
\setlength{\tabcolsep}{3pt}
\begin{tabularx}{\columnwidth}{@{}>{\raggedright\arraybackslash\hangindent=1em\hangafter=1}X l >{\raggedright\arraybackslash}p{0.36\columnwidth}@{}}
\toprule
Quantity & Owner & Rule for pooled execution \\
\midrule
Template contents (\cref{fig:overview}a) & template & stored once, read by all \\
Positions, controls, materials $\boldsymbol\theta_e$ & env. & environment-innermost layout \\
Contacts, Hessian values, contact and MAS factors & env. & no cross-environment terms \\
Barrier stiffness $\kappa_e$, friction lagging & env. & updated from own contacts \\
Newton iterate, energy, line search, termination & env. & no batch-wide test or step \\
PCG scalars $\alpha,\beta$, residual, convergence & env. & segmented reductions, masks \\
CCD bound, capacity overflow & env. & local bound, rollback, replay \\
Physical clock, reset, random stream & env. & addressed by stable identifier \\
\bottomrule
\end{tabularx}
\end{table}

\subsection{Shared templates and environment-innermost storage}
\label{sec:ownership}

We build each batch from a \emph{scene template} (\cref{fig:overview}a). All environments in the batch reference this template; controls, initial states, friction, and material parameters vary through private values, so repeated structure is not copied. Environments with different topology form different batches.

We store fixed-size private arrays with the environment index innermost: scalar component $i$ of environment $e$ lives at address $iB+e$ (\cref{fig:overview}b). This turns repeated topology into memory locality. A warp assigned to one structural item---a tetrahedron, BSR block, or BVH node---reads its indices once from the template and accesses the corresponding values of 32 consecutive environments in coalesced transactions. In the conventional environment-outermost order, the same warp instead gathers items of one environment through repeated index lookups and scattered value accesses. The same layout turns the BSR sparse matrix--vector product into a batched operation: one pass over the shared pattern applies all $B$ operators, as in batched sparse iterative solvers~\citep{aggarwal2021batched}. \Cref{sec:eval-attribution} measures the end-to-end effect of this layout.

Small batches do not fill a warp along the environment axis, so we use several kernel variants---different thread-to-item mappings, blocked reductions, and active-environment compaction paths---selected by static rules when the batch is created. These variants may change the grouping, factoring, or fusion of floating-point operations, but they evaluate the same exact-arithmetic expression (\cref{sec:determinism}).

\begin{algorithm}[t]
\caption{Window execution in \system. All loops over $e$ run in parallel on the GPU; every scalar decision is per environment. \textsc{Advance}$(e)$ sets $k_e\gets k_e{+}1$ and $\mathit{ready}_e\gets\mathit{true}$ and deactivates $e$. As in the reference, a step takes at least two Newton iterations, and a direction that already meets \cref{eq:stopping} takes its CCD-bounded step without an energy test; 88\%, 92\%, and 71\% of the benchmarks' steps take exactly two (\cref{fig:iterations}a). A PCG solve that exhausts its budget returns its last iterate and a line search that exhausts its trials accepts the last one, and $e$ keeps iterating; a step that exhausts any Newton, PCG, or line-search budget is committed when it ends but counted as failed (C2).}
\label{alg:round}
\begin{algorithmic}[1]
\Require template $\mathcal{T}$; states $\mathbf{x}_e$; controls $\mathbf{u}_e(k)$; window lengths $K_e$
\State $k_e\gets 0$, $\mathit{ready}_e\gets\mathit{true}$ for all $e$
\While{$\exists\,e: k_e<K_e$} \Comment{one outer Newton round}
  \ForAll{$e$ with $\mathit{ready}_e$ and $k_e<K_e$} \Comment{refill}
    \State predict $\widetilde{\mathbf{x}}_e$; load $\mathbf{u}_e(k_e)$; lag friction
    \State activate $e$; $\mathit{ready}_e\gets\mathit{false}$
  \EndFor
  \State refit BVHs; find candidates of active $e$ \Comment{per-env.\ roots}
  \State assemble $\mathbf{g}_e,\mathbf{A}_e,\{\mathbf{U}_{e,c}\}$ for active $e$ \Comment{shared pattern}
  \State solve $\mathbf{H}_e\mathbf{d}_e=-\mathbf{g}_e$ for active $e$ \Comment{segmented PCG}
  \ForAll{active $e$}
    \State $\alpha_e\gets$ CCD bound of $e$ along $\mathbf{d}_e$
    \If{$\mathbf{d}_e$ fails \cref{eq:stopping}}
      \State halve $\alpha_e$ until $\Phi_e$ does not increase (${\le}12$ trials)
    \EndIf
    \State $\mathbf{x}_e\gets\mathbf{x}_e+\alpha_e\mathbf{d}_e$; update $\kappa_e$
    \If{$\mathbf{d}_e$ met \cref{eq:stopping} and $e$ iterated twice or more}
      \State \textsc{Advance}$(e)$ \Comment{step completed}
    \ElsIf{the Newton budget of $e$ is exhausted}
      \State report the step as failed (C2); \textsc{Advance}$(e)$
    \EndIf
  \EndFor
  \State roll back and replay every $e$ that overflowed its capacity
\EndWhile
\end{algorithmic}
\end{algorithm}

\subsection{A hybrid operator for changing contacts}
\label{sec:operator}

Contacts are the part of the operator that varies most across environments and Newton iterations. If we assembled every contact term into the sparse matrix, each environment would need a changing per-iteration sparsity pattern, defeating the shared BSR structure. We therefore keep the fixed-pattern terms---inertia, elasticity, joints, and drives---in the shared BSR layout with private values, and apply contact terms as private low-rank factors:
\begin{equation}
 \mathbf{H}_e\,\mathbf{p}=\mathbf{A}_e\,\mathbf{p}+\sum_{c\in\mathcal{C}_e}\mathbf{U}_{e,c}\bigl(\mathbf{U}_{e,c}^{\mathsf T}\mathbf{p}\bigr),
 \label{eq:hybrid-operator}
\end{equation}
where $\mathcal{C}_e$ is environment $e$'s current set of barrier and friction stencils and $\mathbf{U}_{e,c}=\mathbf{J}_c^\mathsf{T}\mathbf{W}_c^{1/2}\in\mathbb{R}^{12\times r}$ factors the stencil Gauss--Newton Hessian $\mathbf{J}_c^\mathsf{T}\mathbf{W}_c\mathbf{J}_c$. Here $\mathbf{J}_c$ maps the stencil's at most four vertices to its $r$ constraint coordinates, and the curvature weight $\mathbf{W}_c$ is clamped positive semidefinite. The rank is one for barriers, because both systems use GIPC's Gauss--Newton barrier Hessian $b''(d)\,\nabla d\,\nabla d^{\mathsf T}$~\citep{huang2024gipc}, and at most two for mollified edge--edge barriers and lagged friction---1.5 on average in our benchmarks rather than a dense $12{\times}12$ block. This factorization represents the same operator as the reference assembly while avoiding a per-iteration sort or merge into a contact-dependent sparsity pattern. For affine bodies, we gather world-space vectors through the affine map and accumulate the products back to that environment's twelve body coordinates.

\begingroup\emergencystretch=1em %
We apply the same ownership split to the MAS preconditioner~\citep{wu2022mas}. The hierarchy and index maps live in the template, while each environment rebuilds its cluster matrices and inverses from its current operator; contact factors contribute $\mathbf{U}\mathbf{U}^{\mathsf T}$ to the relevant cluster blocks exactly as in the single-scene reference.\par
\endgroup

\paragraph{Segmented PCG solves.} We execute PCG for all active environments in shared stages, but keep every recurrence and stopping decision environment-local: inner products, coefficients, residual norms, and convergence flags are reduced per environment, and a converged environment stops updating its iterate. One PCG iteration is fused into five kernels. To avoid paying full-batch cost for a long tail of solves, at batch sizes of at least 256 we compact the remaining environments once at most $\max(64,B/4)$ are still active and finish them on a grid sized to that list.

\subsection{Per-environment collision handling}
\label{sec:collision}

Different environments may occupy identical world coordinates, so collision processing must be segmented just as carefully as PCG. We build a forest of BVHs that shares link structure through the template but refits bounding boxes per environment, and we assign every traversal to one environment at its root; no candidate pair can therefore span two environments. We refit boxes in single precision with outward rounding and confirm candidate pairs with double-precision distance tests, preserving the conservativeness of the reference. We likewise compute the CCD step bound $\alpha_e$ as the minimum over environment $e$'s own pairs.

We also store contact buffers environment-innermost, with one capacity $c$ shared across the batch (\cref{fig:overview}c). Contact demand is heterogeneous, so provisioning every environment for an assumed worst case wastes memory, while truncating a contact list would change the physics and violate C2. If an environment overflows, we roll back only its current step, leave its clock and histories unchanged, grow $c$ between graph launches, rebuild its candidates, and replay the step from the original inputs while the other environments proceed. A task may instead use a preset capacity when a suitable bound is known. The shared capacity keeps addresses affine and coalesced, at the cost of slack and a 32-bit index bound at very large batches; \cref{sec:eval-cost} and Appendix~\ref{app:attribution} quantify the memory/performance tradeoff and the fixed-versus-grown-capacity comparisons.

\subsection{Per-environment Newton control and refill}
\label{sec:scheduling}

We evaluate Newton termination (\cref{eq:stopping}), line search, barrier-stiffness updates, and iteration budgets separately for every environment (\cref{alg:round}). Merely masking converged environments would still impose a common step boundary: every environment would wait for the slowest one before beginning its next step (\cref{fig:design-space}b). Refill removes this barrier. At the start of each outer round, an environment that completed its previous step and has its next input available initializes step $k_e{+}1$, predicts $\widetilde{\mathbf{x}}_e$, reads controls using its own step counter $k_e$, and rejoins assembly; all other environments continue their current Newton solve.

\paragraph{Why refill helps.} Let $N_e$ be the Newton iterations environment $e$ needs for its current step. A lockstep batch spends $\max_e N_e$ Newton rounds on that step, whereas refill follows the environments' individual work. The idealized round-count overhead is therefore
\begin{equation}
  \rho(B)=\mathbb{E}\bigl[\max_{e\le B} N_e\bigr]\,/\,\mathbb{E}[N],
  \label{eq:waste}
\end{equation}
which grows with batch size as the maximum reaches farther into the long tail (\cref{fig:iterations}a). At $B^\star$, $\rho$ ranges from 3.1 to 8.4 across our benchmarks. This does not translate directly into wall-clock speedup because late rounds with few active environments are much cheaper than full rounds; \cref{sec:eval-attribution,sec:eval-hetero} measures the actual refill gain and Appendix~\ref{app:attribution} gives the detailed round accounting.

\paragraph{Choosing the mode.} Refill and lockstep differ only in scheduling: both keep each environment's numerical decisions private, so they implement the same per-environment computation up to floating-point rounding, as formalized in \cref{sec:determinism}. We use refill by default. On highly synchronized workloads, lockstep can be faster because it avoids refill bookkeeping; \cref{sec:eval-hetero} measures this regime.

\paragraph{Remaining synchronization.} Two barriers remain inside a Newton round. PCG must finish the slowest active linear solve before CCD proceeds, and line search waits for environments that backtrack. With a policy in the loop, an environment can also advance only as far as its next action is available: device-resident policies can supply actions without host synchronization, while synchronous collectors bound refill to the action interval (\cref{sec:applications}).

\subsection{Decision-local batching}
\label{sec:determinism}

We use \emph{decision-local batching} as the organizing rule for the nested IPC solver above: data-parallel stages are pooled across environments, while every state-dependent numerical decision remains environment-local. View one environment's solver as a tree of loops---time steps, Newton iterations, line-search trials, and Krylov iterations---whose bodies are data-parallel stages and whose exits are numerical decisions. To preserve the single-environment computation while pooling these stages, we impose four checkable rules:
\begin{description}[leftmargin=2.2em,labelindent=0pt,itemsep=2pt,topsep=3pt]
\item[R0.] A batch-dependent scheduling or kernel choice may change only the grouping, order, factoring, or fusion of arithmetic; for each environment it must evaluate the same exact-arithmetic expression.
\item[R1.] Every mutable value belongs to one environment. A stage may read that environment's state and shared read-only template data, but not another environment's mutable state; inactive environments are not modified, and rollback restores the complete pre-step state.
\item[R2.] Every reduction that produces a value or decision---dot product, energy, CCD bound, convergence test, or accumulation---is segmented by environment.
\item[R3.] Every list produced in parallel, such as collision candidates, contains entries from only one environment per segment, and downstream consumers are insensitive to the order of entries within that segment.
\end{description}
Batch-wide counts and capacities may determine how much work is launched or which R0-equivalent implementation is used, but they do not enter an environment's numerical model or stopping rules.

\begin{proposition}
\label{prop:invariance}
Assume R0--R3, and suppose no step is kept after exhausting its capacity-retry budget and no buffer reaches the index bound of \cref{sec:collision}. In exact arithmetic, after each of environment $e$'s own solver iterations, its state is independent of the batch size, the states and inputs of other environments, and the refill schedule. In floating-point arithmetic, batching can change only the rounding of algebraically equivalent evaluations until such rounding changes one of $e$'s discrete decisions.
\end{proposition}
\begin{proof}[Proof sketch]
Induct over the iterations executed by environment $e$ at every level of its loop tree. Given the same state before an iteration, R1 restricts reads and writes to $e$ and the template, R2 restricts reductions to $e$'s contributions, and R3 prevents generated lists from mixing environments. R0 ensures that any batch-dependent dispatch evaluates the same exact expression. The iteration therefore produces the same values and decisions as a solo run. Inactive rounds leave $e$ unchanged, and rollback erases failed attempts, so changing when the next iteration is scheduled cannot change the exact-arithmetic state.
\end{proof}

The proposition is deliberately not a bitwise guarantee for our default build. Floating-point atomics, fast-math reassociation, batch-size-dependent kernel variants, and PCG compaction can change rounding, and a rounded value can eventually flip a discontinuous decision. We therefore test batch invariance directly in \cref{sec:eval-quality}. We also implement a deterministic mode that fixes contact ordering, accumulation order, reduction trees, kernel variants, and floating-point compilation; it reproduces solo cloth-press runs bitwise across batch sizes, at a substantial throughput cost (Appendix~\ref{app:invariance}). The shared contact capacity is the remaining batch-level resource: the proposition excludes executions that exhaust the retry budget or the 32-bit index bound.

\subsection{Implementation}
\label{sec:implementation}
We implement \system in \kernellang~\citep{hu2019taichi}, in about 31,000 lines of kernels, solver, and scheduler (21,000 in the modules our benchmarks compile), including the reference-equivalent potentials, MAS preconditioner, BVH and CCD, and the batch scheduler. All per-round control is device resident: a window is one CUDA-graph launch whose round loop, line searches, and PCG loops are conditional while-nodes evaluated on the device; active-environment lists are compacted on the device by a single-block scan, and kernels over them run on fixed, saturating grids that read their bounds from device memory, so no grid is resized at run time. The host acts only between launches, to grow contact capacity and replay rolled-back steps. Kernels are compiled once per template and dispatch configuration and cached across processes.

\section{Evaluation}
\label{sec:evaluation}

We evaluate whether \system meets its contract while delivering higher throughput than existing GPU simulators (\cref{sec:eval-throughput,sec:eval-cross,sec:eval-quality}), which design decisions are responsible (\cref{sec:eval-attribution,sec:eval-hetero}), when the approach wins and when it does not (\cref{sec:eval-generality}), and what it costs (\cref{sec:eval-cost}). \Cref{sec:applications} then measures end-to-end application outcomes.

\paragraph{Results overview.} Under the strict contract, \system sustains 36,000, 15,000, and 6,600 environment steps per second on the block press, cloth press, and Franka fold on one H100. The evaluation answers five questions. First, how much faster is the complete IPC system than strong baselines (\cref{sec:eval-throughput,sec:eval-cross})? Second, how much comes from avoiding unnecessary numerical work versus making each remaining iteration cheaper (\cref{sec:eval-throughput,sec:eval-attribution})? Third, does batching preserve each environment's simulation semantics (\cref{sec:eval-quality})? Fourth, which workloads benefit and where do the gains disappear (\cref{sec:eval-hetero,sec:eval-generality})? Finally, \cref{sec:applications} measures whether simulator throughput carries through to policy evaluation, optimization, and learning.

\subsection{Setup}
\label{sec:eval-setup}

\begin{figure*}[t]
\centering
\setlength{\tabcolsep}{1pt}
\begin{tabular}{cccc}
\includegraphics[width=0.23\textwidth]{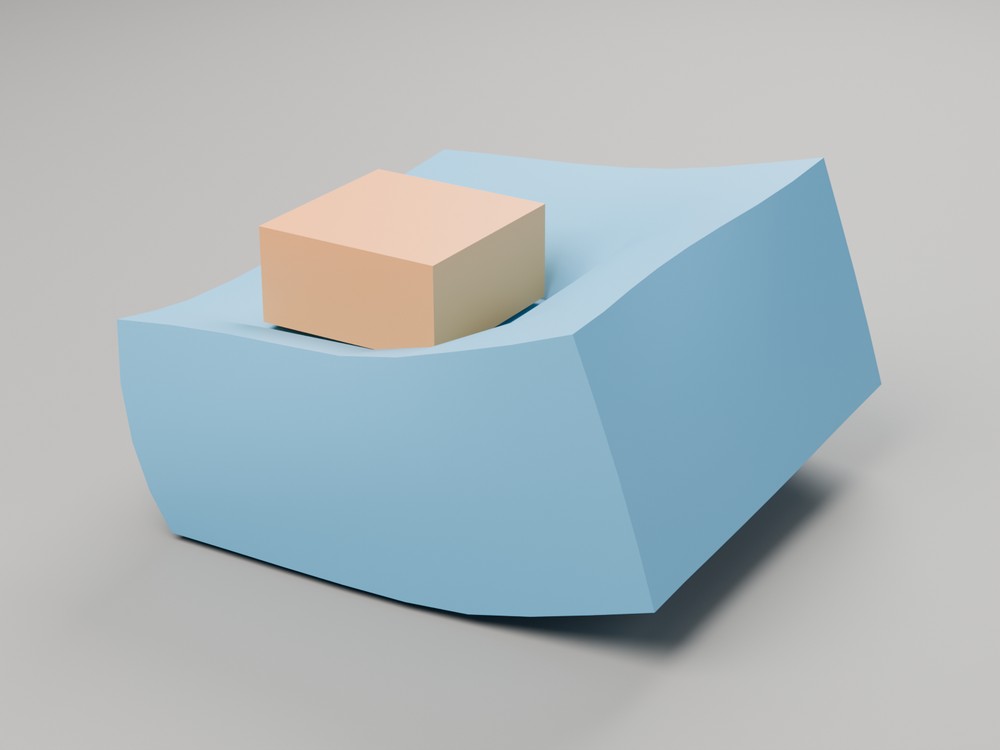} &
\includegraphics[width=0.23\textwidth]{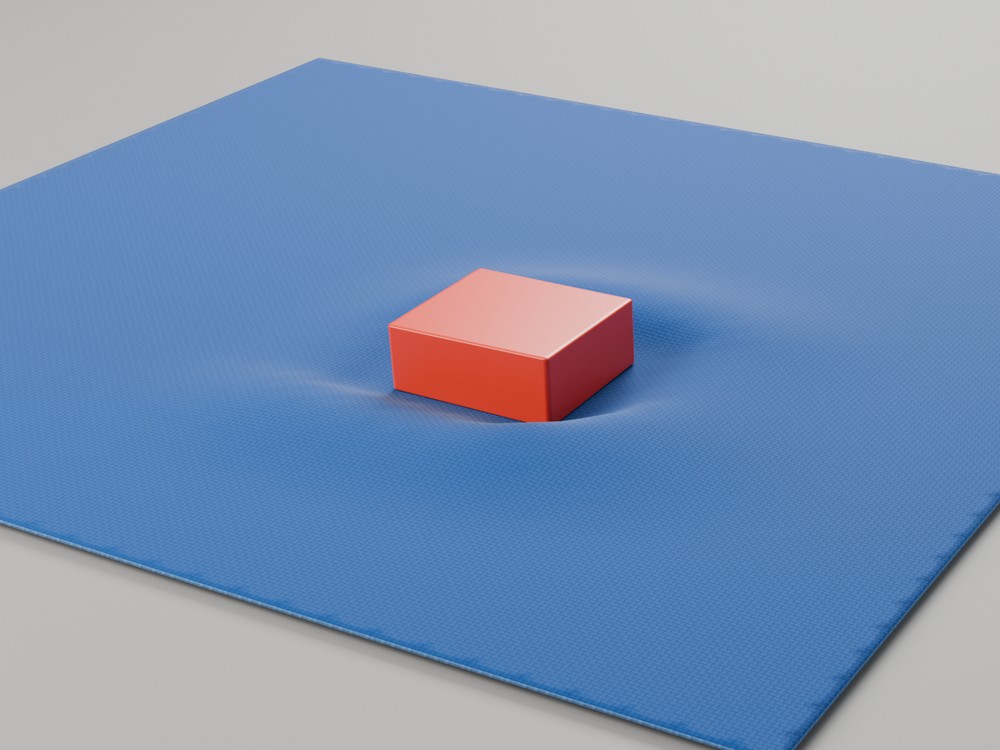} &
\includegraphics[width=0.23\textwidth]{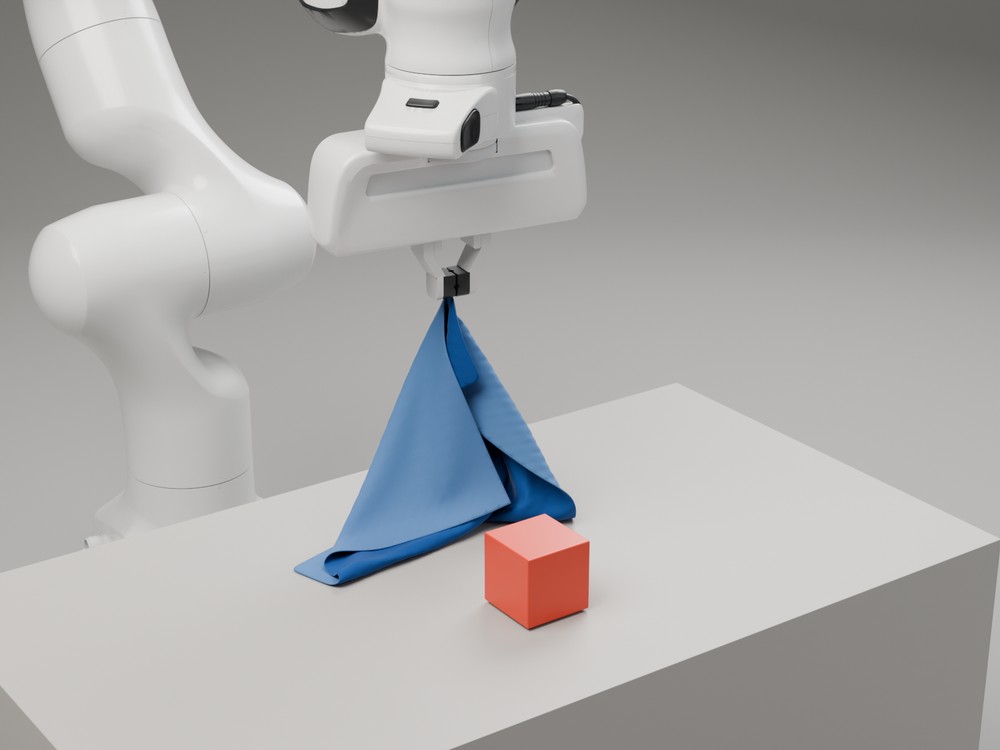} &
\includegraphics[width=0.23\textwidth]{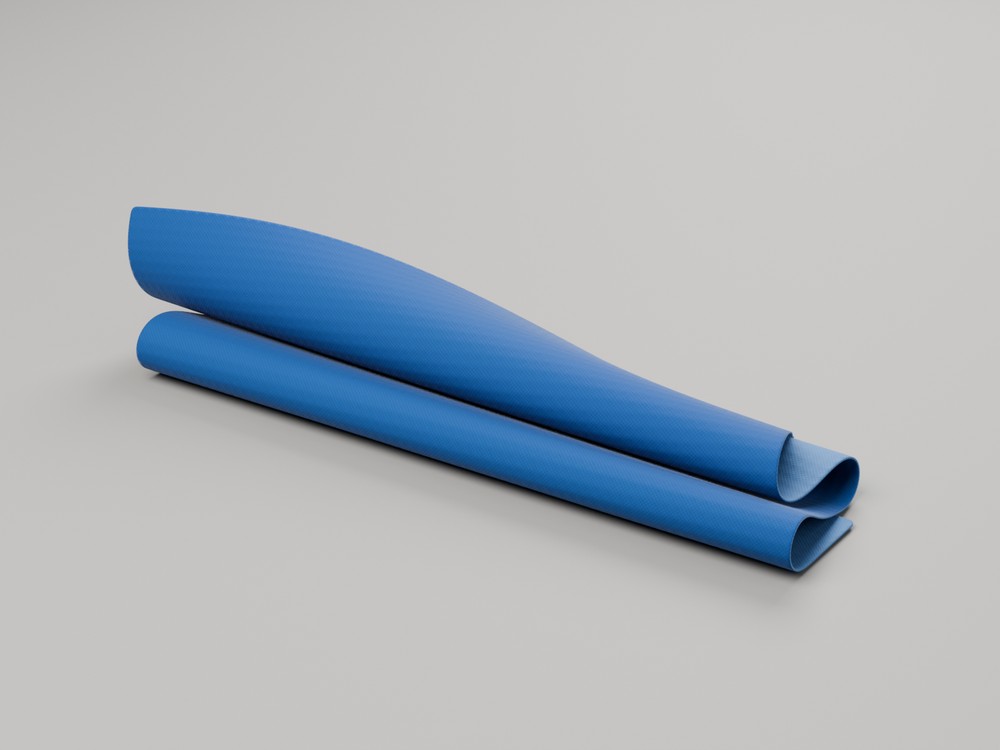} \\
{\sffamily\footnotesize (a) Block press} & {\sffamily\footnotesize (b) Cloth press} &
{\sffamily\footnotesize (c) Franka cloth fold} & {\sffamily\footnotesize (d) Cloth drop (stress test)}
\end{tabular}
\caption{\textbf{Benchmarks span solids, cloth, robots, and self-contact.} Rendered from simulated states. (a) A soft block is pressed, dragged, and released by a soft pusher. (b) A $31{\times}31$ cloth is pressed and dragged on a table. (c) A Franka arm with affine-body links and nine driven joints folds a $25{\times}25$ cloth on a box. (d) A $65{\times}65$ cloth, held vertically, drops onto the ground and folds over itself; all environments land within a few steps of one another. Cloth is drawn with two levels of render-only subdivision and a 0.8\,mm display shell; the drawn surface stays within 1.6\,mm (mean at most 0.23\,mm) of the simulated mesh, whose edges are 5--11\,mm long. The supplementary video shows complete rollouts of all scenes and applications.}
\Description{Four rendered scenes: an orange block on a blue soft block, a small block on a blue cloth, a white robot arm folding a blue cloth on a box, and a large blue cloth lying folded on the ground.}
\label{fig:scenes}
\end{figure*}

\begin{figure*}[t]
\centering
\includegraphics[width=0.92\textwidth]{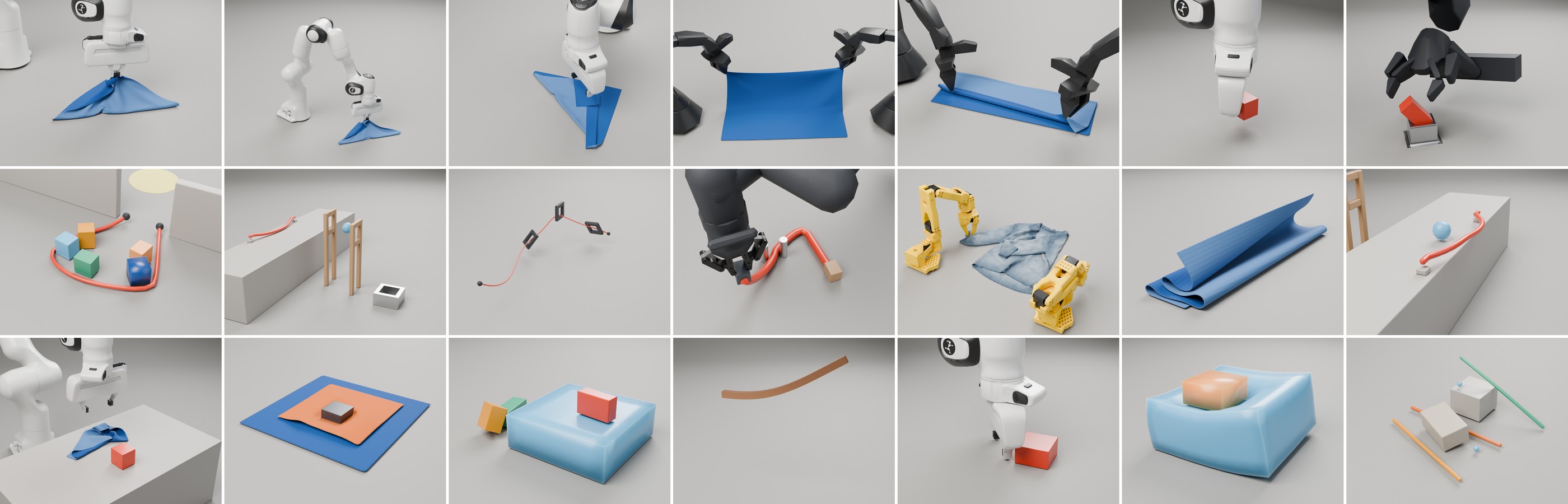}
\caption{\textbf{One system, diverse contact-rich workloads.} Shown are 21 frames from 19 captured rollouts. Top row: cloth folding by three learned policies (two frozen evaluation policies and a checkpoint of the from-scratch PPO run of \cref{sec:applications}), one bimanual towel folding at two instants, cube picking, and peg insertion. Middle row: rope gathering, the narrow-gate slingshot, and wiring through three rings (derivative-free optimization); cable routing (policy evaluation); folding a long-sleeve garment (LeHome); the cloth drop; and the same slingshot rollout at its launch. Bottom row: the Franka fold, two stacked cloths pressed by a block, rigid blocks tumbling onto a soft block, an elastoplastic beam, an imported robot pushing a block, the block press, and codimensional rods and particles dropped onto rigid boxes. All frames are path-traced from simulated states; the supplementary video shows them in motion.}
\Description{A mosaic of 21 rendered simulation frames showing robots manipulating blue cloth, stacked stones, soft blocks, rods, and a bimanual robot folding a towel.}
\label{fig:gallery}
\end{figure*}

\paragraph{Benchmarks.} \Cref{fig:scenes} and \cref{tab:benchmarks} summarize the benchmark scenes, and \cref{fig:gallery} shows the range of scenes \system simulates in this paper, including the reference's codimensional rods and elastoplastic solids, which run through the same pipeline but are not benchmarked separately. Each environment draws its controls---contact onset time, press depth, sliding direction and speed, or the start delay of the arm's scripted fold---from uniform ranges given in Appendix~\ref{app:controls}, so contact onset, sliding, and release occur at different times in different environments, as in real rollout batches. A $65{\times}65$ cloth dropped vertically onto the ground (12,675 DOF, 180 steps) serves as a stress test (\cref{sec:eval-hetero}).

\paragraph{Systems.} \cgq{}\cgqcite{} is a CUDA-graph implementation of the public IPC library libuipc~\citep{libuipc}: it captures the entire Newton loop---assembly, MAS-preconditioned PCG, CCD, and line search---in one CUDA graph with device-side control flow, and with one environment it is 1.6\x and 1.7\x as fast as libuipc on the block and cloth presses (Appendix~\ref{app:libuipc}). It will be released as open source and is the numerical reference that \system reproduces (C3). Its linear solver is a component-partitioned PCG (Appendix~\ref{app:search}). To compare with independently developed systems as well, we include libuipc itself, which descends from GIPC and StiffGIPC~\citep{huang2024gipc,huang2025stiffgipc} and, like \cgq, has no environment axis, and \taccel and Newton (\cref{sec:eval-cross}). Because \cgq has no environment axis, we run it in two configurations. In the \emph{one-scene} configuration, all $B$ environments are tiled side by side in a single scene, which shares launches but makes Newton and line-search decisions for the whole scene; the tiled environments are separate components, so only their PCG solves end independently. In the \emph{multi-process} configuration, $P$ processes, each holding $K$ tiled environments (a $P{\times}K$ \emph{packing}), share the GPU through CUDA MPS. We searched $P$ up to 48 and $K$ in powers of two under the same stopping rule as for \system, 170 configurations of the three benchmarks in all (Appendix~\ref{app:search}); 48 processes do not fit in the H100's memory, and more than 16 did not help.

\paragraph{Protocol.} Throughput is the number of strictly completed environment steps (C2) divided by the wall-clock time of the complete window, including the first step, control upload, capacity recovery, and final synchronization; compilation and initial setup are reported separately (\cref{sec:eval-cost}). For every system, we search batch sizes in powers of two under the same time budget and report each at its own best configuration $B^\star$, and at matched batch sizes for attribution. Comparisons use three interleaved A--B--B--A pairs on the same GPU; we report the median of the paired ratios and, in figures, the observed range (Appendix~\ref{app:throughput} gives the timing window of multi-process configurations). Unless stated otherwise, experiments run on an NVIDIA H100 (80\,GB HBM3) with the precision of \cref{tab:parameters}. The \taccel and Newton comparisons of \cref{sec:eval-cross} run on an NVIDIA RTX PRO 6000 Blackwell (96\,GB), and Appendix~\ref{app:taccel-audit} repeats the \taccel comparison on the H100.

\subsection{Throughput}
\label{sec:eval-throughput}

\begin{figure*}[t]
\centering
\includegraphics[width=\textwidth]{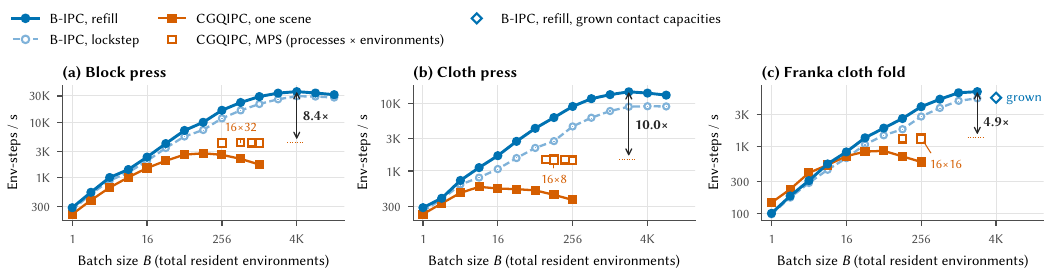}
\caption{\textbf{\system scales to thousands of environments.} Strictly completed environment steps per second versus the total number of resident environments on an H100 (log--log; the ranges of three pairs are smaller than the markers). Squares are the multi-process \cgq configurations timed in three interleaved rounds (\cref{sec:eval-setup}; Appendix~\ref{app:search}), plotted at their total batch; labels name the best one (processes$\times$environments per process). In (c), preset contact capacities do not fit at $B{=}4{,}096$ (\cref{sec:collision}), so that point uses grown ones. Arrows mark the ratio between the best \system configuration and the best \cgq configuration. Refill (solid) is 1.21\x, 1.67\x, and 1.26\x faster than the default build's lockstep path (dashed) at $B^\star$, and at most 1.40\x, 2.24\x, and 1.44\x (at $B{=}256$, 128, and 128).}
\Description{Three log-log plots of throughput against batch size. B-IPC curves rise to a maximum near batch size 2048 to 4096, far above the \cgq{} curves, which saturate before batch size 128. Single markers show multi-process configurations of \cgq{} and StiffGIPC.}
\label{fig:scaling}
\end{figure*}

\Cref{fig:scaling} plots throughput against batch size. \system scales over three orders of magnitude of batch size before the GPU saturates, at $B^\star$ between 2,048 and 4,096 environments; beyond $B^\star$, throughput declines slightly as the window's drain and the memory traffic of the largest batches grow faster than utilization. \cgq saturates much earlier: its one-scene configuration peaks at 8--128 environments, because a tiled scene iterates as long as its hardest environment, and 16 MPS processes raise its ceiling by 1.6--2.6\x, still far below \system. At each system's best configuration, \system is 8.4\x faster on the block press, 10.0\x on the cloth press, and 4.9\x on the Franka fold (\cref{tab:benchmarks}); against \cgq's best single-process (one-scene) configuration, it is 13.1\x, 25.7\x, and 7.6\x faster. Tightened until its final-state error does not exceed that of \cgq's best packings ($\varepsilon_v/5$), \system is still 5.9\x and 5.0\x faster than them on the block and cloth presses (Appendix~\ref{app:throughput}). The clean contribution of private progress is visible within \system itself: refill is 1.21\x, 1.67\x, and 1.26\x faster than the default build's lockstep path at $B^\star$. Lockstep execution keeps everything else of \system---one process, the environment axis with our layout, shared templates, the hybrid operator, PCG compaction, and per-environment convergence masks---but restores a common step boundary, and it is already 6.9\x, 6.0\x, and 3.9\x faster than \cgq's best configuration (the best-vs-best speedup divided by the refill gain). An environment-outermost build of \system is 1.29\x slower on both presses (\cref{sec:eval-attribution}).

\begin{table*}[t]
\caption{Benchmarks and strictly completed environment steps per second (env-steps/s) on an H100 at each system's best configuration $B^\star$ (multi-process: processes $\times$ environments per process). Speedups over \cgq{}\,/\,libuipc are medians of paired ratios, not ratios of the medians shown (Franka fold: 4.87\x against 6,562/1,356${}={}$4.84\x); we did not port the Franka fold to libuipc. DOF are per environment and exclude the driven pusher (24~DOF); the Franka fold's 12 affine bodies are the arm's links, the table, and a riser. Models: stable Neo-Hookean~\citep{smith2018snh}, Baraff--Witkin~\citep{baraff1998cloth}.}
\label{tab:benchmarks}
\centering\small
\setlength{\tabcolsep}{3.0pt}
\begin{tabular}{@{}llrrrrrrrr@{}}
\toprule
Scene & Model & DOF & $h$ (s) & Steps & $B^\star$ & \system & \cgq ($B^\star$) & libuipc ($B^\star$) & \begin{tabular}[b]{@{}r@{}}Speedup over\\\cgq{}\,/\,libuipc\end{tabular} \\
\midrule
Block press & 1,536-tet stable Neo-Hookean & 1,215 & 1/60 & 360 & 4,096 & 35,862 & 4,294 (16$\times$32) & 4,369 (16$\times$32) & 8.4\x{}\,/\,8.2\x \\
Cloth press & 1,800-tri.\ Baraff--Witkin cloth & 2,883 & 1/60 & 360 & 2,048 & 14,960 & 1,495 (16$\times$8) & 2,056 (16$\times$8) & 10.0\x{}\,/\,7.3\x \\
Franka cloth fold & 12 affine bodies, 9 joints, 1,152-tri.\ cloth & 2,019 & 1/100 & 1,130 & 2,048 & 6,562 & 1,356 (16$\times$16) & -- & 4.9\x \\
\bottomrule
\end{tabular}
\end{table*}

\paragraph{Work avoided.} Throughput factors into Newton iterations per environment step, PCG iterations per Newton iteration, and wall time per environment-PCG iteration (Appendix~\ref{app:throughput}). \system reduces the first factor substantially: relative to \cgq's best packings, it needs 1.55\x, 1.9\x, and 1.65\x fewer Newton iterations on the block press, cloth press, and Franka fold because no environment waits for a scene-wide convergence test. Per Newton iteration, the two systems require essentially the same PCG work on the block press and Franka fold and \system uses 1.34\x fewer PCG iterations on the cloth press. Overall, per-environment decisions save 1.55--2.5\x of the PCG iterations per environment step, almost entirely through fewer Newton iterations.

\paragraph{Work made cheaper.} The larger factor is the cost of the work that remains: each environment-PCG iteration is 5.4\x, 3.9\x, and 2.9\x cheaper in \system. This is a batching effect rather than a uniformly faster single-environment implementation: at $B{=}1$, \system is 1.31\x, 1.25\x, and 0.69\x as fast as \cgq on the three benchmarks. At scale, one wide batch shares index tables and matrix structure and executes the solver stages together; on identical synchronized block-press inputs, \system issues 74.9\% fewer warp instructions and moves 75.5\% less DRAM traffic per step. \Cref{sec:eval-attribution} separates the contributions of the layout, shared templates, contact representation, refill, and compaction, while Appendix~\ref{app:throughput} gives matched-width and per-stage profiles.

\paragraph{Per-environment decisions alone are not enough.} To isolate this point, we modified \cgq so that every environment in a tiled scene has its own Newton and PCG stopping decisions and line search. The modification matches \system's Newton iteration counts closely (2.23, 2.39, and 2.49 iterations per environment step versus 2.23, 2.38, and 2.56), but improves \cgq by only 1.05\x, 1.20\x, and 1.09\x. \system remains 8.0\x, 8.5\x, and 4.5\x faster because the tiled scene still detects contact, assembles, and builds its preconditioner for all environments until the last one converges. The main result therefore comes from making independent numerical progress compatible with one wide, structure-sharing execution, not from changing the stopping test alone. Appendix~\ref{app:attribution} describes the modification and its trajectory checks.

\subsection{Comparison with other GPU simulators}
\label{sec:eval-cross}

\begin{figure}[t]
\centering
\includegraphics[width=\columnwidth]{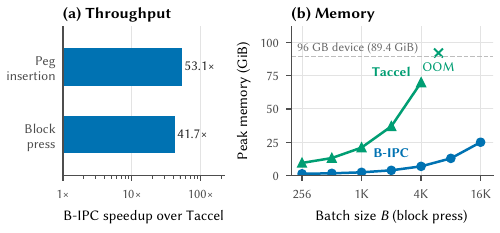}
\caption{\textbf{Faster and more memory-efficient than \taccel.} RTX PRO 6000. (a) Median paired throughput ratios at each system's best batch size, under the peg insertion's original response gates (Appendix~\ref{app:taccel-audit}) and the block press's per-system checks. (b) Peak memory on the block press: \taccel exhausts the GPU at $B{=}6{,}144$, whereas \system uses 12.8\,GiB at $B{=}8{,}192$.}
\Description{A bar chart with speedups of 53 and 42 times over Taccel on the peg insertion and the block press, and a plot of memory against batch size in which Taccel rises steeply and B-IPC stays low.}
\label{fig:cross}
\end{figure}

\begin{figure*}[t]
\centering
\includegraphics[width=\textwidth]{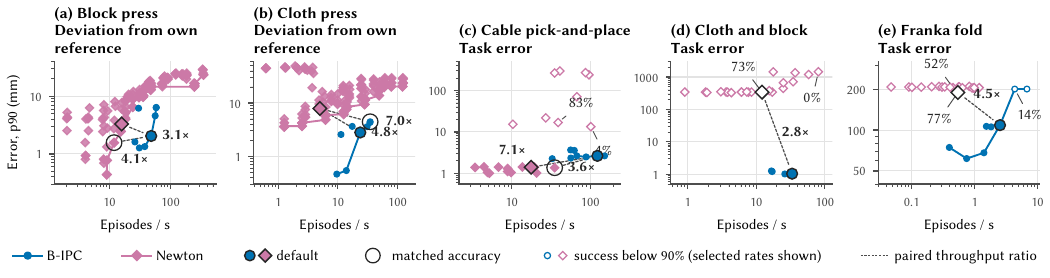}
\caption{\textbf{Faster than Newton at comparable accuracy.} RTX PRO 6000: error (90th percentile over 128 held-out environments) against throughput at each system's best batch size; lower and further right is better. On the presses, the error is the deviation from each system's own reference solution: \system at a quarter of its time step; for VBD, (a) its converged solution, and (b), where VBD does not converge in its iteration count, four substeps of 200 iterations. Appendix~\ref{app:vbd} shows the presses against the IPC reference; (c--e) show task errors, defined with success in Appendix~\ref{app:vbd}. Circles are \system's configurations, with the tolerance sweep at its full time step connected; diamonds are Newton's. On the presses, a line connects Newton's non-dominated configurations. Large markers are the defaults, rings the matched-accuracy configurations, and open markers configurations that succeed in fewer than 90\% of the environments (selected rates are given). Dashed lines join paired configurations and carry the median throughput ratio of three interleaved pairs. The vertical ranges differ between panels.}
\Description{Five log-log scatter plots of error against episodes per second. On the two presses, where each system is measured against its own reference solution, Newton's most accurate points reach about half a millimeter on the block press and 3.6 millimeters on the cloth press, B-IPC's points lie between about half a millimeter and 6.4 millimeters, and Newton's fastest points extend further right at errors above 13 millimeters. On the cable scene Newton's successful points lie lower and B-IPC's lie further right. On the cloth and block scene and on the fold most Newton points are open markers, meaning low success.}
\label{fig:newton}
\end{figure*}

\paragraph{Comparison with libuipc.} libuipc is a public single-scene GPU IPC library with MAS-preconditioned PCG. We port the block and cloth presses to its Python interface with the same geometry, materials, controls, $\hat d$, friction, Newton and PCG tolerances, line-search budget, and preconditioner, and batch it as a practitioner would: environments tiled into one scene, several worlds in one process, and from 2 to 48 processes under CUDA MPS, 118 configurations in all. Only steps that libuipc reports as converged within its budgets count, and every final state passes its intersection check. Its best configurations, 16 processes of 32 tiled block presses and of 8 cloth presses, reach 4,369 and 2,056 env-steps/s, 1.02\x and 1.38\x \cgq's best, and \system is 8.2\x and 7.3\x faster (\cref{tab:benchmarks}; all 118 configurations in \cref{tab:search-libuipc} of Appendix~\ref{app:search}). With both systems at libuipc's own looser defaults ($\varepsilon_v{=}0.05$\,m/s, PCG tolerance $10^{-3}$, block-Jacobi preconditioning, each at its best configuration), \system remains 3.7\x and 6.0\x faster. Appendix~\ref{app:libuipc} gives implementation differences, single-environment rates, final states, and the reasons for the lower ratios.

\paragraph{Comparison with \taccel.} \taccel is an independently developed Warp IPC system with affine bodies and per-environment CG reductions. We use it as a complete-system comparison, not as an isolation of our batching design: the systems also differ in preconditioner, precision choices, and implementation. On \taccel's peg-insertion benchmark and on the block press, at each system's best batch size on an RTX PRO 6000, \system completes 53.1\x and 41.7\x more environment steps per second (\cref{fig:cross}). The peg insertion passes the benchmark's original cross-system response gates. On the block press, that gate is not meaningful at large batches because \taccel's batched trajectories differ from its own single-environment trajectories by 8.5--21\,mm; at $B{=}1$ the systems agree within the 5\,mm gate, while \system's batched and single-environment trajectories agree within 0.05\,mm. We therefore report the block press using each system's own completion, tracking, and inversion checks at matched Newton tolerance. Memory also limits \taccel to smaller batches: it reaches 70\,GiB at $B{=}4{,}096$ and runs out of memory at $B{=}6{,}144$, whereas \system holds $B{=}8{,}192$ in under a fifth of that footprint (\cref{fig:cross}b). Appendix~\ref{app:taccel-audit} gives the full gate, preconditioner, and H100 audit.

\paragraph{Comparison with Newton.} NVIDIA Newton~\citep{newtonphysics} uses VBD~\citep{chen2024vbd} for deformables and STEEL for articulated bodies, with penalty-style contact and single precision. Because its model and solver differ from IPC, a raw throughput ratio at one setting would confound numerical accuracy with execution cost. We therefore compare error--throughput curves on five scenes with randomized inputs: the two presses, a cable pick-and-place, a block carried in a cloth hammock, and the Franka cloth fold (\cref{fig:newton}). Each system is evaluated against its own converged or refined reference where possible, and we compare defaults as well as matched-accuracy configurations; Appendix~\ref{app:vbd} gives the complete configuration grids and reference checks.

On the presses, \system's defaults are both faster and closer to their own references than Newton's defaults: 3.1\x and 4.8\x faster at 2.0 and 2.8\,mm error versus 3.3 and 7.9\,mm for VBD. VBD can be faster when coarse agreement is acceptable, overtaking \system's fastest configurations around 14--15\,mm error. The robot examples show the same tradeoff rather than a universal winner. Newton is more accurate on the cable (1.4\,mm versus 2.6\,mm), while \system is 7.1\x faster at the defaults and 3.6\x faster than Newton's cheapest configuration that matches \system's default accuracy. On the cloth-and-block scene and the fold, \system is 2.8\x and 4.5\x faster at the defaults and succeeds in 100\% and 98\% of environments, versus 73\% and 77\% for Newton's defaults. These are complete-system comparisons; the contact formulation, precision, and articulated/deformable coupling differ along with batching.

\subsection{Quality and batch invariance}
\label{sec:eval-quality}

We test the three parts of the contract on the default build, with the candidate-reach filters enabled at every batch size (\cref{sec:determinism}).

\begin{figure}[t]
\centering
\includegraphics[width=\columnwidth]{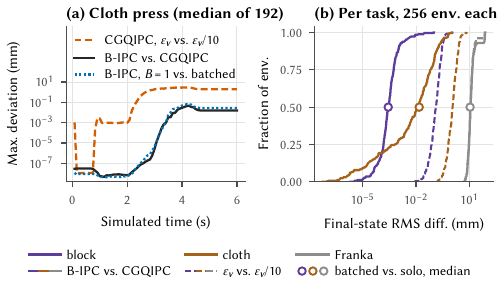}
\caption{\textbf{\system reproduces the reference within the reference's own tolerance sensitivity.} (a) Largest vertex deviation over the cloth press, median over 192 probes at each sampled step (every five steps), between \system and \cgq (black), between \cgq runs at tolerance $\varepsilon_v$ and $\varepsilon_v/10$ (orange), and between a probe run alone and in batches of 64 and 4,096 with randomized, reset, and deactivated neighbors (blue). (b) Distributions of final-state RMS differences per task (environments 0--255 of each task; the probes of (a) and of C1 are 0--191); on the Franka fold, the dashed curve coincides with the solid one; open circles mark the median difference between a probe run alone and in a batch of 4,096 (C1).}
\Description{A log plot over time in which the batched-versus-solo and B-IPC-versus-\cgq{} deviations stay far below the \cgq{} tolerance deviation, and cumulative distributions in which B-IPC-versus-\cgq{} differences are mostly smaller than tolerance differences.}
\label{fig:quality}
\end{figure}

\paragraph{Batch invariance (C1).} For each task, we select 192 probe environments and run each at $B\in\{1,64,4{,}096\}$ with the other environments re-randomized, reset at random steps, or deactivated, and once more at $B{=}1$ in a separate process, all from preset capacities (at $B{=}4{,}096$ the Franka fold's contact-term preset is halved to fit the index range of \cref{sec:collision}). With unfixed rounding (\cref{sec:determinism}), none of the 1,728 probe runs, not even the second $B{=}1$ run, is bitwise identical to its first $B{=}1$ run. We detect no effect of batch composition beyond this rounding effect (two-sample tests over the 192 probes, $p\ge0.30$; Appendix~\ref{app:invariance} gives the tails). The median final-state RMS difference from a probe's $B{=}1$ run is 0.27\,\textmu{}m, 14.5\,\textmu{}m, and 11\,mm on the three tasks at $B{=}4{,}096$; at $B{=}64$ and in a second $B{=}1$ run it is 0.29--0.32\,\textmu{}m, 14.9--15.1\,\textmu{}m, and 11--12\,mm. Tightening \cgq's tolerance tenfold changes the same 192 environments by 0.13, 1.13, and 12\,mm. On the presses, batching therefore perturbs trajectories by 0.2\% and 1.3\% of the reference's own tolerance sensitivity (\cref{fig:quality}a shows the cloth-press trace). The Franka fold is chaotic: rounding alone changes the final fold by 92\% as much as the tenfold tolerance change. At the level of task outcomes, batching changes no more than rerunning does: between a solo run and the same episode inside a batch of $B^\star$ with re-randomized and reset neighbors, success flips in 3.0--3.1\% of 2,048 pick-cube episodes, 5.7--6.9\% of 1,024 peg insertions, and 0--0.4\% of all 256 cable-routing episodes of \cref{sec:applications}, against 3.7\%, 6.0\%, and 0.4\% between two solo runs. The differences in flip rate are equivalent to zero within a margin of $\pm2$ percentage points (Appendix~\ref{app:invariance}).

Final states of a chaotic task cannot reveal a coupling smaller than the rounding spread, so we also compare per iteration. With neighbors that start in extreme states, are re-randomized and deactivated before every Newton round, or hold NaN and Inf values, no discrete decision of any probe changes on the presses over 20 steps, and the iterates differ from those of an unperturbed batch by at most 1.5\,nm, against 0.9\,nm between two unperturbed runs; Appendix~\ref{app:invariance} describes this poison test and a replicate test. In the deterministic mode of \cref{sec:determinism}, the cloth press is bitwise reproducible: every probe matches its solo run at all 360 steps in batches of 64, 512, 2,048, and 4,096 (refill and lockstep execution), and two runs of a full 4,096-environment batch match each other at every step (Appendix~\ref{app:invariance}).

\begin{figure}[t]
\centering
\includegraphics[width=0.495\columnwidth]{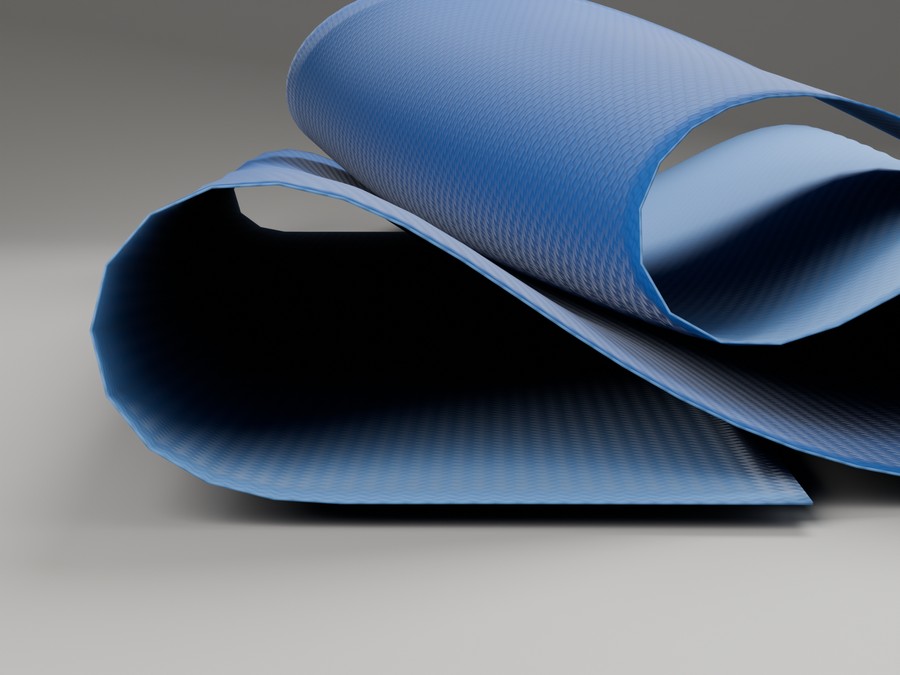}\hfill
\includegraphics[width=0.495\columnwidth]{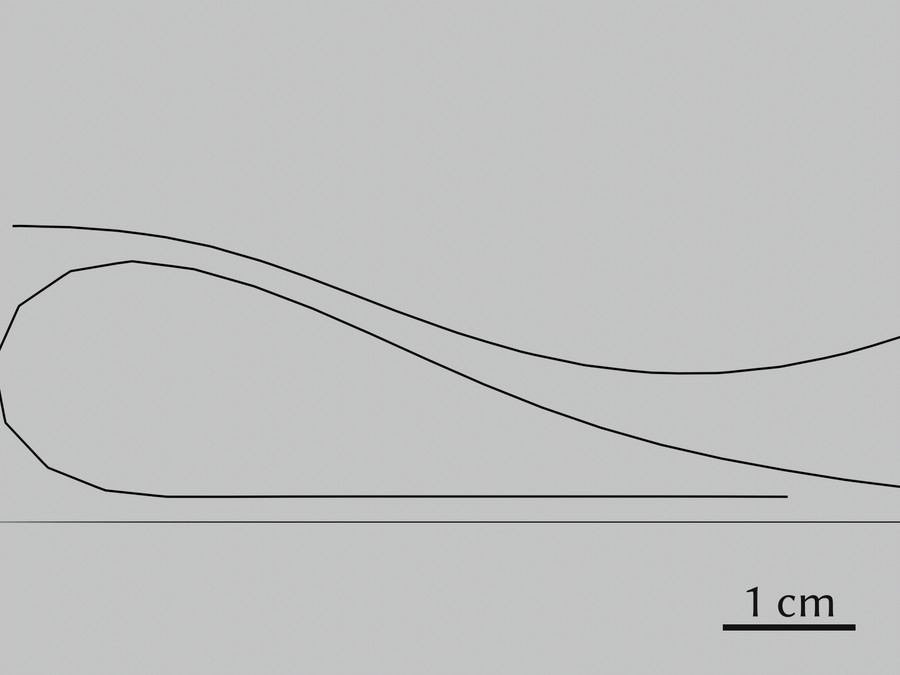}
\caption{\textbf{IPC preserves separation under layered self-contact.} Cloth drop drawn from the raw simulated triangles. Left: the folded cloth at $t{=}0.6$\,s. Right: a section along the roll axis at $t{=}3$\,s (scale bar 1\,cm). At every sampled time, the layers come no closer than 1.85\,mm, inside the 2\,mm barrier distance, and never intersect.}
\Description{Left, a close-up render of a blue cloth folded over itself in two layers. Right, a thin cross-section of the same cloth showing the layers as separated curves, with a one-centimeter scale bar.}
\label{fig:selfcontact}
\end{figure}

\paragraph{Strict completion (C2).} No step fails on the block and cloth presses for either system; on the Franka fold, both systems report the few steps that exhaust the line-search budget, commit them without a retry, and exclude them from the counted throughput; counting them as completed changes the speedup by 0.02\%. We check intersection-freeness with a separate global test that shares no code with the solver's collision detection. On 2.5 million stored states (the 768 common environments at every fifth step, one full batch of each benchmark at $B^\star$, and every held-out scene), an exact edge--triangle test over all pairs of the contact model finds no intersection, and the smallest distance over all point--triangle and edge--edge pairs is never below the smallest barrier-pair distance that the solver records for the step that starts from that state. The smallest barrier-band pair distance over all steps and Newton iterates is 0.11$\hat d$ for both systems on the 768 environments, and the layers of the self-contacting cloth drop stay at least 1.85\,mm apart (\cref{fig:selfcontact}).

\paragraph{Reference equivalence (C3).} Because \system and \cgq accumulate floating-point sums in different orders, which contact-rich trajectories amplify, the question is whether their differences exceed those that the solver tolerance permits anyway; \cref{fig:quality} compares them with \cgq run at its tolerance and at a ten times tighter one. Across environments 0--255 of each task (768 in all), the median final-state difference between \system and \cgq is 0.27\,\textmu{}m, 12\,\textmu{}m, and 11\,mm on the three tasks, whereas the tolerance alone induces 0.13, 1.1, and 12\,mm in \cgq; \system is closer to \cgq than \cgq at $\varepsilon_v$ is to \cgq at $\varepsilon_v/10$ in 648 of the 768 environments (Appendix~\ref{app:invariance} accounts for the other 120); batch composition and the difference to the reference are of the same size in every task.

\subsection{Where does the speedup come from?}
\label{sec:eval-attribution}

\begin{figure*}[t]
\centering
\includegraphics[width=\textwidth]{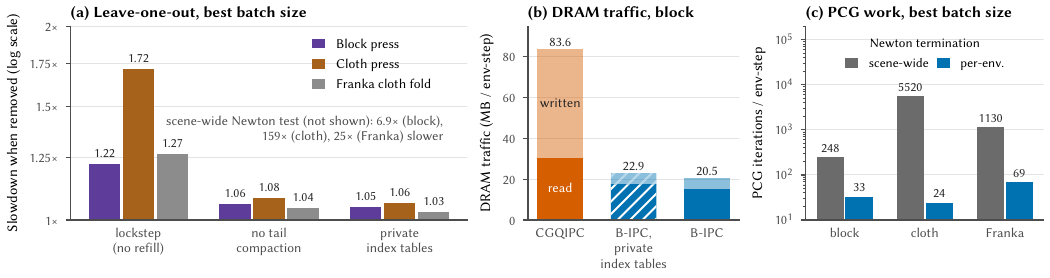}
\caption{\textbf{Private progress avoids work; shared structure makes the rest cheaper.} (a) Slowdown when one component is removed from \system at $B^\star$, everything else kept: refill (replaced by a lockstep barrier in the same kernel), compaction of the PCG tail, and sharing of the template's index tables (replaced by private copies); median of three blocks of fresh runs. Replacing per-environment Newton termination by a scene-wide test is far slower and is noted in the panel; (c) shows its PCG work. Each bar is a separate leave-one-out comparison. (b) DRAM bytes read and written per environment step on identical synchronized block-press inputs ($B{=}128$; \cgq in its one-scene configuration; hardware counters); the hatched bar is \system with private index tables for every environment. (c) PCG iterations per environment step with scene-wide and per-environment Newton termination at $B^\star$.}
\Description{Grouped log-scale bars of slowdown factors for four removed components on three tasks, largest for scene-wide Newton termination; stacked bars of read and write traffic; and log-scale bars of PCG iterations per step with scene-wide and per-environment termination.}
\label{fig:ablation}
\end{figure*}

Removing one component at a time from \system at $B^\star$ (\cref{fig:ablation}a) shows where the throughput comes from.

\paragraph{Per-environment Newton control.} Of the components removed from \system, it matters most. Without it, a batch must iterate until every environment passes the stopping test \eqref{eq:stopping} in the same iteration, so its Newton iteration count follows the slowest environment: at $B^\star$ that alone is 4.1\x, 8.4\x, and 3.1\x the per-environment count on the block press, cloth press, and Franka fold ($\rho$ of \cref{eq:waste}, from per-step logs). If environments that have already converged keep taking Newton steps under their own line search, as in our ablation, some of those steps overshoot the tolerance and fail the test again, and the stacked test then waits far longer: throughput falls to 0.145\x, 0.0063\x, and 0.040\x of \system on the block press, cloth press, and Franka fold (\cref{fig:ablation}c; Appendix~\ref{app:attribution}). This arm is a pessimistic upper bound for a batched solver without per-environment control; the fair arm keeps per-environment termination and makes converged environments wait at a lockstep barrier in the same kernel, which costs 1.22--1.72\x (below). The saving is work avoided, not work made cheaper.

\paragraph{Shared templates and environment-innermost storage.} These reduce the cost of the work that remains. On identical synchronized inputs, which remove all differences in Newton work, \system issues 74.9\% fewer warp instructions than \cgq and moves 75.5\% less DRAM traffic per step, reading 50.5\% fewer bytes and writing 89.7\% fewer (\cref{fig:ablation}b, $B{=}128$). Giving every environment private copies of the template's index tables costs 3--6\% of throughput and adds 9--13\% DRAM traffic at $B^\star$ (hatched bar: 12\% on the block press at $B{=}128$). We measure the environment-innermost layout end to end against an environment-outermost build of \system from the same source: private arrays are stored per environment, adjacent threads process adjacent items of one environment, and every layout-specific choice is made in that build's favor (Appendix~\ref{app:attribution}). At $B^\star$ it is 1.29\x slower on both presses (1.293\x and 1.286\x; three interleaved pairs), with the same Newton and PCG iteration counts. It loses in three stages: assembling the Hessian, assembling the preconditioner's cluster matrices, and the broad phase of collision detection each take 1.7--2.1\x as long; its PCG stage is not slower. The low-rank contact operator is measured with kernel microbenchmarks on the same $B^\star$ states. Per Newton iteration, the low-rank contact factors are 2.8\x and 2.0\x cheaper than the cheapest assembled alternative, an unmerged list of dense per-stencil $12{\times}12$ blocks as GIPC-lineage solvers keep it. Assembled blocks merged into sorted per-environment patterns apply faster but must be re-sorted every Newton iteration, which pays off only above about 50 PCG iterations per Newton iteration; the benchmarks average 10--27 (Appendix~\ref{app:attribution}).

\paragraph{Refill and compaction.} In the leave-one-out blocks, replacing refill by a lockstep barrier in the same kernel costs 1.22\x, 1.72\x, and 1.27\x at $B^\star$, within 3\% of the default build's separate lockstep path (1.21--1.67\x, \cref{sec:eval-throughput}), and compaction of the PCG tail adds 4--8\%. After all optimizations, PCG accounts for 44--47\% of \system's time on the block press and no other stage exceeds a quarter (Appendix~\ref{app:attribution}), which is why every stage must be batched.

\begin{figure}[t]
\centering
\includegraphics[width=0.9\columnwidth]{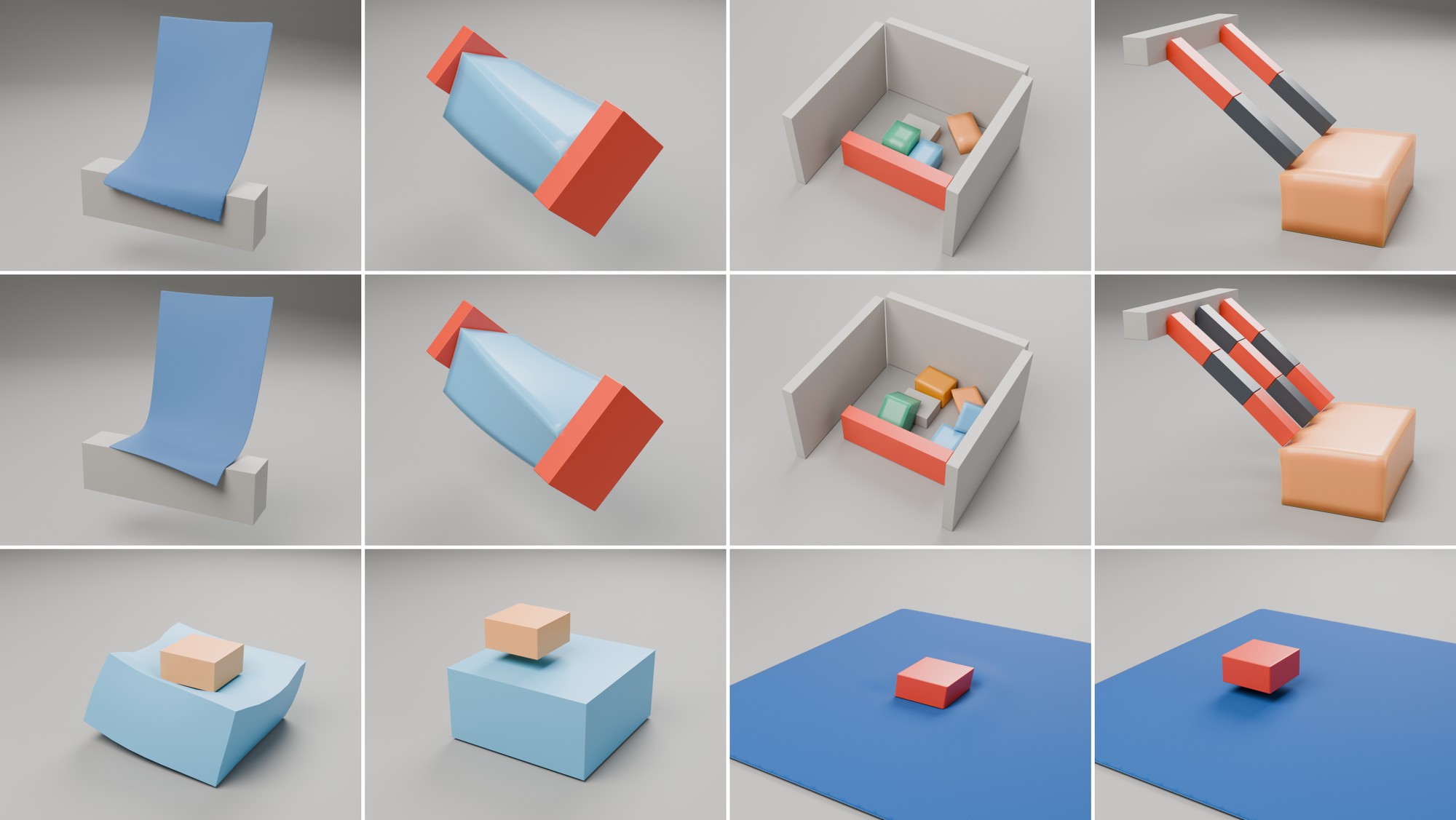}
\caption{\textbf{Held-out scenes test generality.} Rendered held-out scenes and periodic-contact benchmark. Columns: cloth pulled around a bar, a tetrahedral strip compressed and twisted by affine clamps, soft objects in a container, and an articulated affine-body mechanism pressing a pad; top row small and middle row large variants, each showing environment 0 at its main contact. Bottom row: the periodic block and cloth presses (\cref{sec:eval-hetero}), two environments of one batch at the same instant, one pressed and sliding and one lifted.}
\label{fig:heldout}
\end{figure}

\begin{figure*}[t]
\centering
\includegraphics[width=\textwidth]{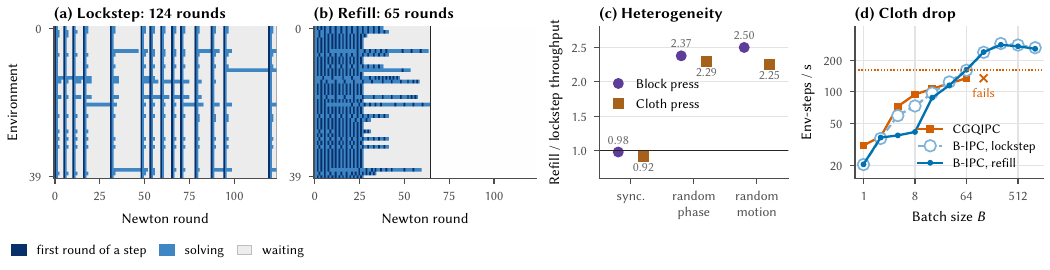}
\caption{\textbf{Refill exploits heterogeneous progress.} (a, b) Activity of 40 block-press environments that each complete the same window of physical steps, under lockstep execution and under Newton-round refill (dark blue: first round of a step; blue: solving; gray: waiting). Refill completes the window in 65 instead of 124 rounds; the remaining waiting is the drain at the end of the window. (c) Measured refill/lockstep throughput ratio at $B{=}256$ over 720 steps as contact phases change from synchronized to randomized phases or motion amplitudes. (d) The cloth drop, where lockstep execution is faster at small batches and marginally faster at $B^\star$; the dotted line marks the best \cgq{} multi-process configuration, and the cross marks where a single \cgq{} scene fails.}
\Description{Two raster plots of environment activity over Newton rounds; the lockstep raster has many gray waiting cells, the refill raster is dense and ends earlier. A third plot shows ratios near 1 for synchronized contact and about 2.3 to 2.5 for randomized contact. A fourth plot shows throughput of the cloth cloth drop against batch size.}
\label{fig:hetero}
\end{figure*}

\paragraph{The single-environment gap.} At $B{=}1$, \system is 1.31\x and 1.25\x faster than \cgq on the block and cloth presses but 0.69\x as fast on the Franka fold and 0.66\x on the cloth drop: the default build's kernels are tuned for large batches, and its small-batch variants only change thread mappings. The best-configuration speedups of \cref{tab:benchmarks} therefore do not come from a faster single-environment solver but from how the batch is organized.

\paragraph{Why IPC benefits.} The individual mechanisms are not unique to IPC, but their value depends on the solver. Per-environment termination alone adds only 1.05\nobreakdash--1.20\x to \cgq's tiled scenes because the remaining stages still execute for the whole scene (\cref{sec:eval-throughput}). IPC benefits much more from the full organization for two reasons: its Newton work varies strongly between environments (\cref{fig:iterations}), and its sparse stages are memory intensive, so sharing read-only structure and laying private values environment-innermost substantially reduce execution cost. These effects reinforce one another, yielding 4.9--10.0\x on the benchmarks and 5.2--15.4\x on the held-out scenes (\cref{sec:eval-generality}). Whether the same combination pays off in another solver depends on its iteration variability, shared structure, and memory behavior.

\subsection{Heterogeneity and refill}
\label{sec:eval-hetero}

Refill pays off only if environments progress at different rates. We isolate this dependence with a periodic-contact benchmark in which every environment presses, slides, and releases repeatedly (\cref{fig:heldout}, bottom), and vary only how contact phases are distributed across the batch (\cref{fig:hetero}c). When all environments contact in synchrony, refill brings nothing and its bookkeeping costs 2\% on the block and 8\% on the cloth. With randomized contact phases, refill is 2.37\x and 2.29\x faster than lockstep execution; with randomized motion amplitudes, 2.50\x and 2.25\x (Appendix~\ref{app:regime} matches the modes' contact-buffer capacity). These gains exceed those at $B^\star$ on the main benchmarks (1.21--1.67\x over the default build's lockstep path) because the periodic benchmark repeats contact throughout the window. The activity rasters in \cref{fig:hetero}a,b show the mechanism.

\paragraph{A stress test.} The cloth drop combines conditions under which our design gains little (\cref{fig:hetero}d): its environments all land within a few steps of one another, so refill has little variation to exploit; each has 12,675 degrees of freedom, so a few dozen environments already occupy much of the GPU; and once landed, each cloth folds onto itself with over a thousand self-contact pairs (Appendix~\ref{app:regime}). Here refill runs at 0.97\x the throughput of lockstep at $B{=}256$, and in two independent sets of paired runs \system is 1.73\x (refill) and 1.77\x (lockstep) faster than the best \cgq configuration (8 MPS processes $\times$ 16 environments), whose single scene fails beyond $B{=}64$ (Appendix~\ref{app:regime}).

\subsection{Generality and regime}
\label{sec:eval-generality}

\begin{figure*}[t]
\centering
\includegraphics[width=\textwidth]{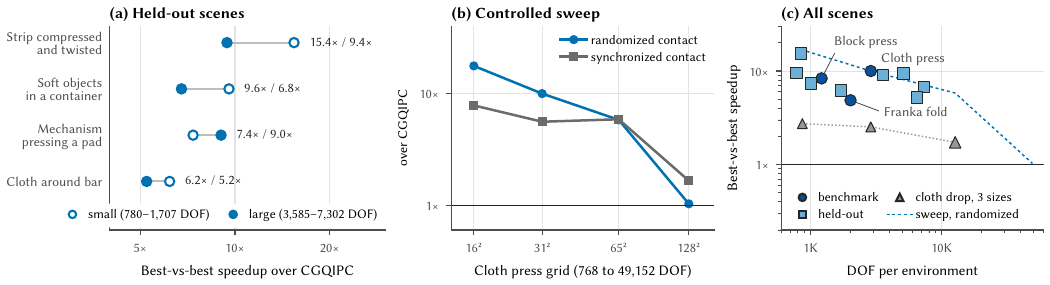}
\caption{\textbf{Largest gains: moderate-size, heterogeneous environments.} (a) Held-out scenes: best-vs-best throughput ratios over \cgq for the small and large variant of each family (median of three pairs; ranges smaller than markers). (b) Controlled cloth-press sweep at four resolutions with randomized or synchronized contact phases; lines show speedup over \cgq. (c) Best-vs-best speedup of every scene against its size; triangles mark the cloth drop at three resolutions.}
\Description{Left, a dot plot of held-out speedups. Middle, two speedup curves over \cgqplain{} across cloth resolutions, with randomized contact above synchronized contact at small sizes and both falling with size. Right, a scatter of all scenes showing the same size trend.}
\label{fig:generality}
\end{figure*}

All execution thresholds of \system were tuned on the benchmarks above. To test whether its gains depend on that tuning, we generated eight held-out scenes in four families---cloth pulled around a bar, a tetrahedral strip compressed and twisted by affine clamps, three or six soft objects in a container, and an articulated affine-body mechanism pressing a pad---each at a small (780--1,707 DOF) and a large (3,585--7,302 DOF) scale (\cref{fig:heldout}), with self-contact and per-environment control variation over 320 steps. Scenes and parameters were fixed before any measurement. \system runs all eight from one capacity preset that was not chosen per scene, with growth enabled; one scene outgrows it once, and the retried steps are inside its timed window. With all dispatch rules and thresholds frozen, and each system at its best configuration under the protocol of \cref{sec:eval-setup}, \system outperforms \cgq on all eight: by 6.2--15.4\x on the small and 5.2--9.4\x on the large variants (\cref{fig:generality}a).

\paragraph{When does \system win?} The controlled cloth sweep separates environment size from heterogeneity (\cref{fig:generality}b). With randomized contact, the speedup over \cgq falls from 17.7\x at 768 DOF to 10.0\x at 2,883 DOF and 5.8\x at 12,675 DOF, and reaches 1.0\x at 49,152 DOF, where one environment already occupies the GPU and the fine cloth carries about 129,000 active self-contact pairs per environment. Synchronized contact still gains 5.6--7.8\x up to 12,675 DOF, because synchronized contact phases do not make the environments' Newton and PCG work identical; heterogeneity adds a further 1.8--2.3\x at the smaller resolutions but essentially nothing at 12,675 DOF. The held-out scenes follow the same trend: the small variants gain 6.2--15.4\x and the large variants 5.2--9.4\x.

Self-contact-rich, PCG-bound workloads gain less even at the same mesh size. The $65^2$ cloth press gains 5.8--5.9\x, while dropping the same-resolution cloth gains only 1.8\x: after landing, the drop needs about 92 PCG iterations per Newton iteration instead of 12--13 on the press, and an iteration is only 1.5\x cheaper in \system instead of 5.4\x (Appendix~\ref{app:regime}). The same sweep also reinforces the complementary tradeoff with VBD: VBD can be attractive when coarse error is acceptable, but its most accurate configuration reaches \system's default accuracy in only one of the eight size/phase cells; Appendix~\ref{app:vbd} gives the full curves. A practical rule of thumb is that \system pays off most---roughly 6--18\x in our tests---for deformable environments of up to a few thousand degrees of freedom that progress at different rates, and still about 5--9\x up to roughly ten thousand DOF. The advantage shrinks as a single environment fills the GPU or the solve becomes dominated by per-environment self-contact work.

\subsection{Memory and startup}
\label{sec:eval-cost}
Memory grows linearly with batch size. At $B{=}2{,}048$, the Franka fold allocates 21.4\,GiB (22.3\,GiB device peak), of which the shared template is 1.2\,MiB, private state 4.8\,GiB, and contact capacity 16.6\,GiB at the task's preset capacities; private copies of the template would add 2.5\,GiB. Growing from small capacities reduces the Franka contact storage to 4.0\,GiB, but 0.65\% of step attempts are replayed during growth and throughput falls to 0.79\x the fixed-capacity mode; the corresponding throughput ratios are 0.99\x and 0.70\x on the block and cloth presses. Appendix~\ref{app:attribution} gives the matched capacity-mode comparisons. Kernels are compiled once per scene template: compiling the Franka fold's kernels takes about three minutes (166--186\,s), and with every kernel cached across processes a run starts in 8\,s from process launch to the first timed step.

\section{Applications}
\label{sec:applications}

\begin{figure*}[t]
\centering
\setlength{\tabcolsep}{1pt}
\begin{tabular}{ccc}
\includegraphics[width=0.305\textwidth]{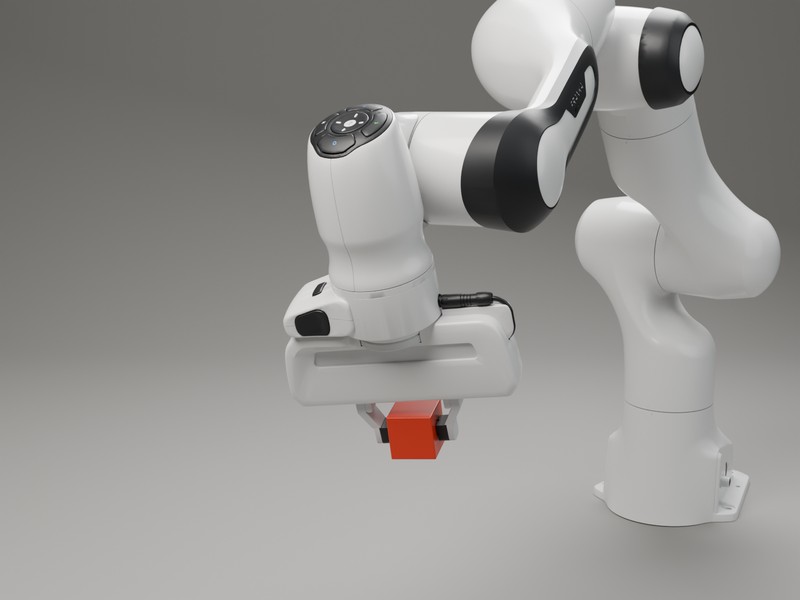} &
\includegraphics[width=0.305\textwidth]{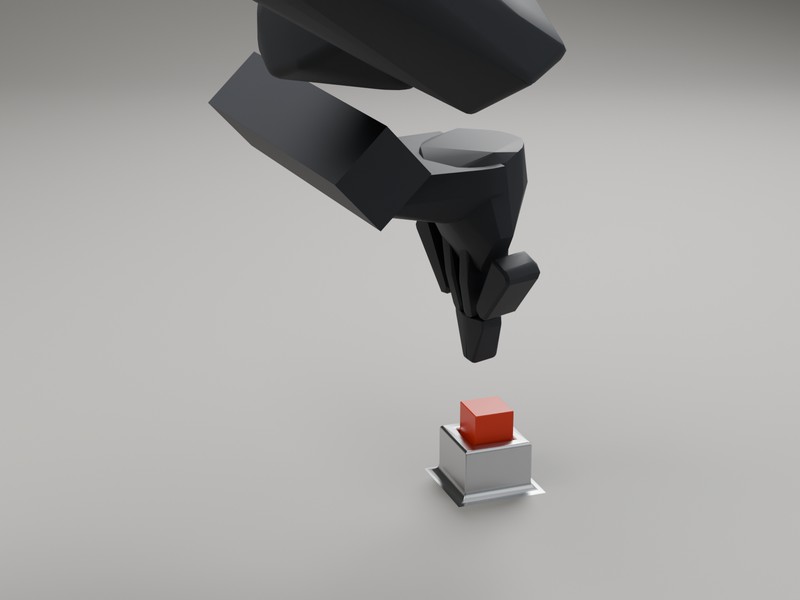} &
\includegraphics[width=0.305\textwidth]{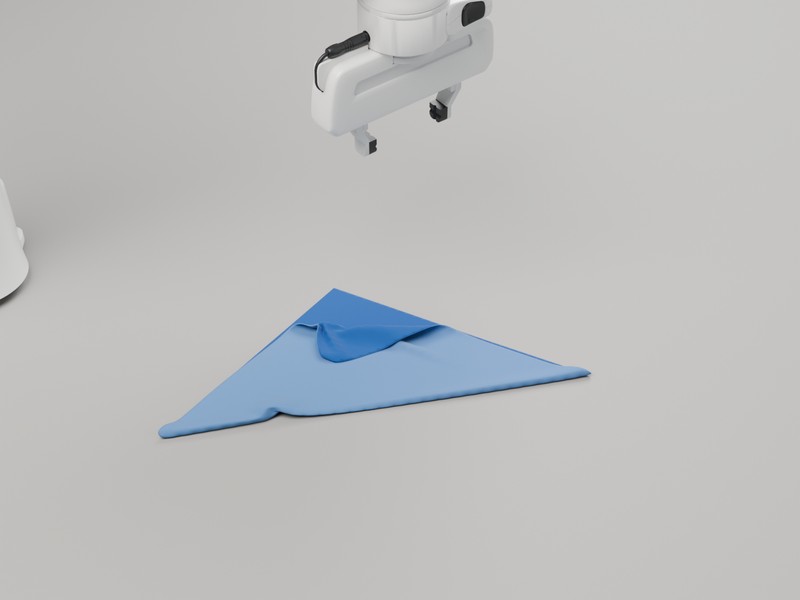} \\[-1pt]
{\sffamily\scriptsize pick cube (evaluation)} & {\sffamily\scriptsize peg insertion (evaluation)} & {\sffamily\scriptsize cloth fold (PPO)} \\[2pt]
\includegraphics[width=0.305\textwidth]{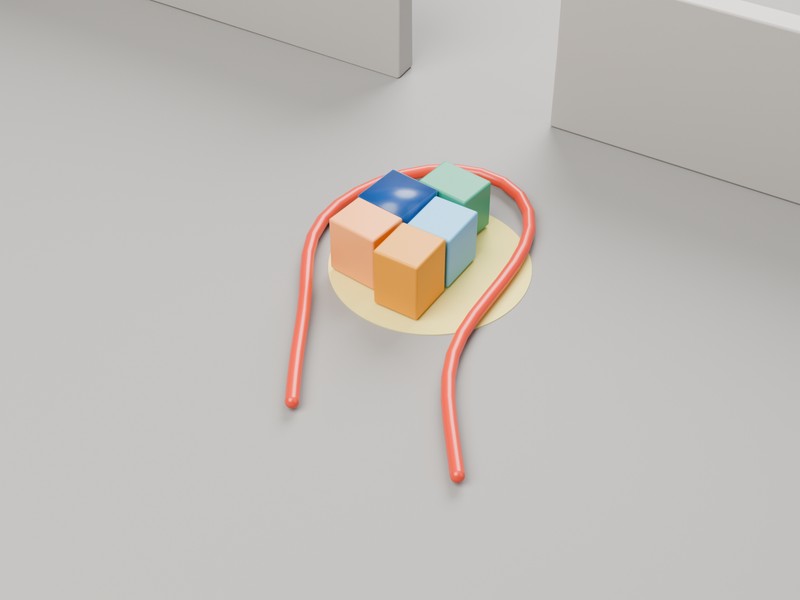} &
\includegraphics[width=0.305\textwidth]{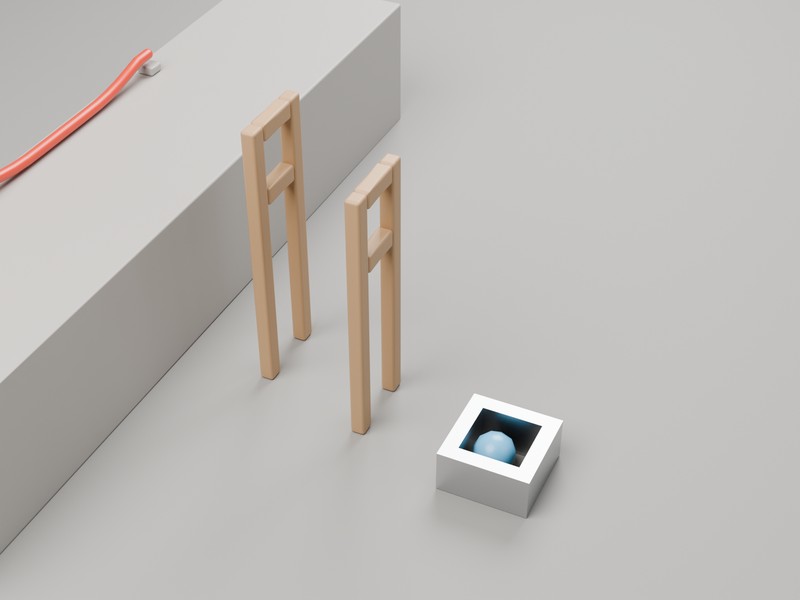} &
\includegraphics[width=0.305\textwidth]{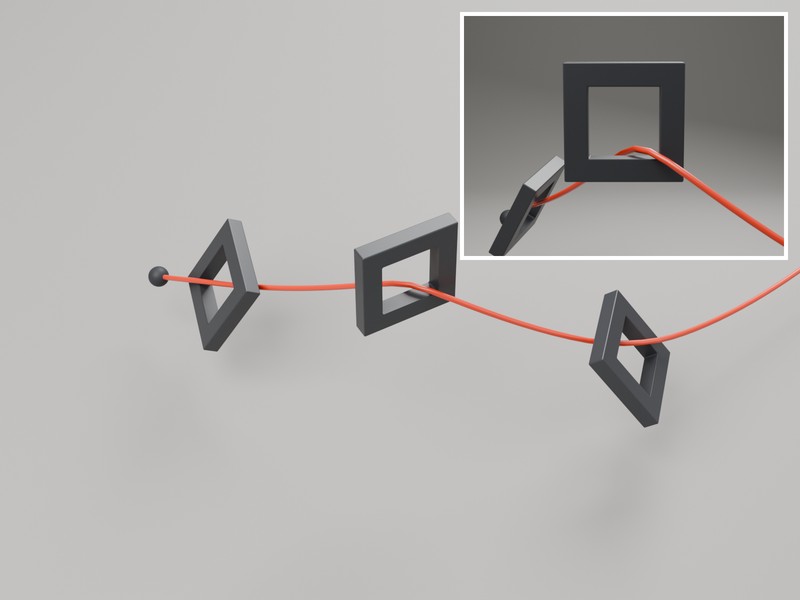} \\[-1pt]
{\sffamily\scriptsize rope gathering (optimization)} & {\sffamily\scriptsize narrow-gate slingshot (optimization)} & {\sffamily\scriptsize 3D ring wiring (optimization)}
\end{tabular}\\[2pt]
\includegraphics[width=\textwidth]{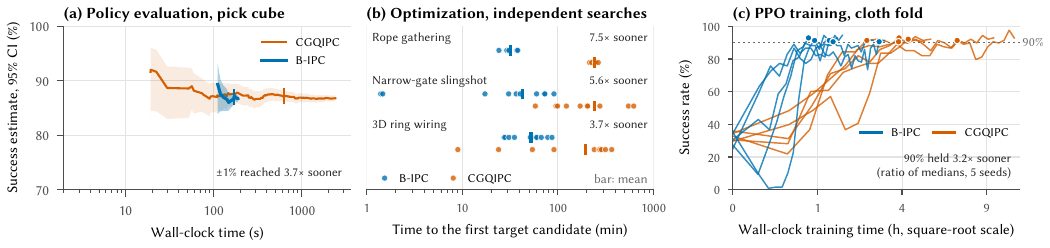}
\caption{\textbf{End-to-end gains when physics matters.} Top: final states of representative episodes, rendered from simulated states; the three optimization panels show the first candidate found by the search. Bottom: (a) estimate of a frozen policy's success rate with its 95\% Wilson interval over the first $n$ dispatched episodes, plotted when all $n$ have finished; ticks mark when the interval first narrows to $\pm1\%$. (b) Time to the first target candidate of every independent search (dots; 10 per task and backend, 5 on the rope gathering, with the same seeds and the same population of 2,048 candidates per generation on both backends) and its mean per task and backend (bars); the labels give the ratios of the means. (c) Success rate of PPO on the cloth fold on 128 evaluation cloths, one line per seed (five per backend); dots mark the first evaluation at or above 90\%, and each run stops after three consecutive evaluations at or above 90\%.}
\Description{Six rendered manipulation scenes above three plots. In each plot, the B-IPC curve reaches the target several times earlier than the \cgq{} curve.}
\label{fig:applications}
\end{figure*}

Simulation throughput is a means; what a user wants is an evaluated policy, an optimized motion, or a trained controller. We therefore measure three applications end to end. In each, \system and \cgq run the same task and algorithm with the same settings and quality gates on the same GPU; \cgq uses its fastest \emph{healthy} configuration, one whose screening runs finish without native errors or stalls. Wall-clock time includes policy inference, observation and reward computation, resets, and optimizer updates. The applications use each task's native iteration budgets rather than the strict C2 contract of \cref{sec:evaluation}; both backends use the same task-level budgets and treatment of capped steps, and a capacity overflow is replayed rather than accepted. Cap and overflow events are rare except in the cloth-fold PPO workload discussed below; Appendix~\ref{app:applications} reports their frequencies and retry protocol. Unless stated otherwise, timings exclude one-time start-up on both backends; Appendix~\ref{app:applications} also reports warm- and cold-start results. \Cref{fig:applications} shows the scenes and curves, and \cref{tab:applications} summarizes the tasks.

\paragraph{Policy evaluation.} We first evaluate three frozen policies with inexpensive observations: a PPO policy that picks up a cube~\citep{tao2025maniskill3} (16,384 episodes), a state expert for peg insertion (4,096 episodes), and a SARNN cable-routing policy~\citep{ichiwara2022sarnn,murooka2026robomanipbaselines} with a $64{\times}64$ rasterized camera (256 episodes, four runs per backend). Success rates agree within 1.1 percentage points, while \system evaluates 6.4--12.7\x more episodes per second (geometric mean 8.7\x; \cref{tab:applications}). Asynchronous completion changes which episodes finish first---failed pick-cube episodes take roughly twice as much Newton work---so an unbiased success estimate must wait for every episode in a fixed set. Under that criterion, \system narrows the pick-cube success interval to $\pm1\%$ 3.7\x sooner (\cref{fig:applications}a). Refill helps when the policy can cheaply serve arbitrary subsets of environments: it is 1.23--1.53\x faster than lockstep on pick cube and peg insertion, but slower on cable routing because rendering tiny active subsets is inefficient.

\begin{figure}[t]
\centering
\includegraphics[width=\columnwidth]{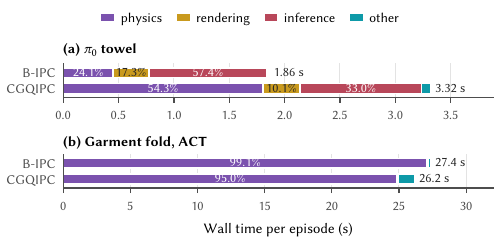}
\caption{\looseness=-1 \textbf{Non-physics work bounds end-to-end gains.} Time breakdown of the two visual policy evaluations, in seconds per episode on one RTX PRO 6000 (medians of the three timed runs of \cref{tab:applications}); each bar adds up to the inverse of that table's throughput, and the labels give shares of it. (a) $\pi_0$ towel: \system in lockstep at $B{=}384$, \cgq with 16 environments tiled in one scene; physics is \system's batched advance and \cgq's scene step. (b) Garment fold: \system with refill at $B{=}128$, \cgq as 16 single-environment processes under MPS; both render only the environments that request an image. \system's physics is here its wall time minus the timed stages, i.e., its solver rounds plus refill scheduling; \cgq's physics is its scene step, and its shares are means over its 16 processes, with each process's idle time under ``other''. Rendering is the policy cameras' path tracer, inference the policy network ($\pi_0$ server call, ACT forward pass).}
\Description{Two panels of horizontal stacked bars, B-IPC above \cgqplain. Towel: B-IPC's bar is 1.86 s, mostly inference and rendering; \cgqplain's is 3.32 s, more than half physics. Garment fold: B-IPC 27.4 s and \cgqplain{} 26.2 s, both almost entirely physics.}
\label{fig:app_stages}
\vspace{-6pt} %
\end{figure}

\paragraph{When physics is not the bottleneck.} Two visual evaluations show where simulator speedup stops carrying through. For a bimanual towel task with a visual $\pi_0$ policy~\citep{zhao2023aloha,black2024pi0}, rendering and network inference take 75\% of \system's time; its physics is 4.0\x faster, but the complete application is only 1.8\x faster (\cref{fig:app_stages}). At the other extreme, LeHome's long-sleeve garment~\citep{li2026lehome} has 43,632 degrees of freedom per environment, close to the regime where one environment already fills the GPU. There \cgq is 1.05\x faster end to end, consistent with the size sweep of \cref{sec:eval-generality}. Appendix~\ref{app:applications} gives the native-simulator quality comparisons; \cref{fig:app_stages} gives the detailed stage accounting.

\begin{table}[t]
\caption{End-to-end application results on an H100; the $\pi_0$ towel and garment-fold rows run on an RTX PRO 6000. Policy-evaluation entries are throughput (episodes/s); optimization and learning entries are time to target (min and h). Task quality is given per row (\system{}\,/\,\cgq{}). Rates are ranges over three paired runs; single values are medians, except optimization times, which are means over seeds. Gain is the median paired throughput ratio (\system{}\,/\,\cgq{}), except for learning, where it is the ratio of median times (\cgq{}\,/\,\system{}) over five seeds (95\% bootstrap intervals: pick cube 5.1--9.0\x, cloth fold 1.4--7.5\x), and optimization, where it is the ratio of mean times (intervals in the text). Header means are geometric. Steps/act.\ is simulation steps per policy action (none for open-loop plans); R/L is refill throughput over lockstep within \system. Sustained means three consecutive evaluations at or above 90\%. $^a$Includes charged native errors and stalls. $^b$Ten seeds per task, five on the rope gathering.}
\label{tab:applications}
\centering\footnotesize
\setlength{\tabcolsep}{2.2pt}
\begin{tabular}{@{}lrrrrr@{}}
\toprule
Task {\scriptsize(quality \system{}\,/\,\cgq{})} & \system & \cgq & Gain & Steps/act. & R/L \\
\midrule
\multicolumn{6}{@{}l}{\emph{Evaluation throughput, no path tracing} (episodes/s; geometric mean 8.7\x)} \\
Pick cube {\scriptsize(86.8\%\,/\,86.9\%)} & 84--91 & 6.8--7.0 & 12.7\x & 5 & 1.53 \\
Peg insertion {\scriptsize(96.0\%\,/\,97.1\%)} & 12.2--13.1 & 1.95 & 6.4\x & 12 & 1.23 \\
Cable routing {\scriptsize(94.1\%\,/\,93.9\%)} & 5.38 & 0.670--0.674 & 8.0\x & 48 & 0.46 \\
\midrule
\multicolumn{6}{@{}l}{\emph{Evaluation throughput, path-traced cameras} (episodes/s)} \\
ALOHA towel, $\pi_0$ {\scriptsize(2.3\%\,/\,3.5\%)} & 0.533--0.540 & 0.300--0.302 & 1.8\x & 25 & n/a \\
Garment fold, ACT {\scriptsize(61.7\%\,/\,58.9\%)} & 0.036--0.037 & 0.038--0.039 & 0.95\x & 100 & 1.09 \\
\midrule
\multicolumn{6}{@{}l}{\emph{Optimization time to target} (min; mean over seeds$^b$)} \\
Rope gathering & 31.9 & 239.8$^a$ & 7.5\x & -- & -- \\
Narrow-gate slingshot & 42.5 & 239.8$^a$ & 5.6\x & -- & -- \\
3D ring wiring & 52.5 & 193.0$^a$ & 3.7\x & -- & -- \\
\midrule
\multicolumn{6}{@{}l}{\emph{Learning time to target} (PPO, h to sustained 90\%; 5 seeds; geometric mean 4.7\x)} \\
Pick cube & 0.52 & 3.56 & 6.9\x & 5 & 1.40 \\
Cloth fold & 1.45 & 4.65 & 3.2\x & 3 & 1.00 \\
\bottomrule
\end{tabular}
\vspace{-6pt} %
\end{table}

\begingroup\emergencystretch=1em
\paragraph{Derivative-free trajectory optimization.}
Contact-rich manipulation objectives are discontinuous in the control parameters, so we optimize three open-loop manipulation tasks with CEM~\citep{rubinstein1999cem} or CMA-ES~\citep{hansen2001cmaes}: rope gathering, a narrow-gate slingshot, and three-ring wiring. Both backends use the same optimizer, seeds, population of 2,048 candidates, and task gates; \system evaluates one generation as a batch, while \cgq uses its fastest healthy multi-process configuration. \system evaluates candidates 11.6\x, 2.9\x, and 3.0\x faster, and over paired independent searches reaches the target 7.5\x, 5.6\x, and 3.7\x sooner on the three tasks (\cref{fig:applications}b, \cref{tab:applications}).

Time-to-target need not equal candidate throughput because small numerical differences can change candidate rankings and hence how many generations a stochastic search needs. The ring-wiring searches require essentially the same number of generations, while the slingshot and rope searches do not. The rope exposes the important fairness case: \cgq's scene-wide stopping test oversolves most candidates and shifts their guide losses. Tightening \system's tolerance tenfold reproduces those losses; even under this matched-convergence setting, five searches with the same seeds reach the target 6.8\x sooner than \cgq's. Thus the application gain remains large after matching the numerical effect responsible for the ranking shift. Appendix~\ref{app:applications} reports the generation statistics, loss comparisons, native-error accounting, and retry protocol.
\par\endgroup

\paragraph{Reinforcement learning.} We train PPO~\citep{schulman2017ppo} on pick cube from scratch and on a cloth-folding task in which a Franka arm grasps, carries, folds, and releases a $31{\times}31$ cloth (\cref{fig:teaser}); the latter starts from a policy with a pretrained arm controller. Both backends use identical hyperparameters, seeds, and evaluations. On pick cube, the policies require similar numbers of updates, so training time tracks sample throughput: the complete \system loop processes samples 7.8\x faster and reaches sustained 90\% success a median 6.9\x sooner over five seeds (95\% bootstrap interval 5.1--9.0\x). Exploratory policies create more varied contact than the converged evaluation policy, so the training collector benefits more from per-environment decisions and refill.

On cloth fold, physics is 96\% of \system's training loop. \cgq's best configuration uses 32 environments in one tiled scene and averages 4.3 Newton iterations per step, versus 2.2 for \system; 38\% of its steps hit the task's Newton cap with the scene-wide test unmet, versus 2.1\% of \system's environment steps. The complete \system loop processes samples 3.8\x faster, while refill itself contributes essentially nothing because an action spans only three simulation steps. Across five seeds, \system reaches sustained 90\% success a median 3.2\x sooner (95\% interval 1.4--7.5\x; \cref{fig:applications}c). Cross-evaluation on the same episodes gives comparable final policy success on the two simulators (Appendix~\ref{app:applications}), so the measured training gains reflect faster data generation rather than a systematically easier evaluation backend.

\section{Discussion and Limitations}
\label{sec:discussion}

\paragraph{When does sharing structure pay off?} \system targets batches of environments with repeated topology, moderate size, and heterogeneous progress---the regime of robot rollouts (\cref{fig:generality}b). Its advantage shrinks as each property weakens: environments large enough to occupy the GPU on their own leave little to pool---at 49,152 degrees of freedom per environment, \system is 1.0--1.7\x as fast as \cgq's best configuration---synchronized contact leaves nothing for refill to exploit, and PCG-bound, self-contact-rich scenes such as the cloth drop gain 1.8--2.7\x: there an iteration costs 1.5\x less than \cgq's, against 5.4\x on the press of the same mesh. The execution mode, kernel variant, and batch size are currently chosen by static rules or by the user; selecting them automatically from measured throughput would recover the few percent that refill costs on synchronized scenes and whatever static kernel selection leaves on unseen scenes.

\paragraph{Topology and capacity.} A batch shares one template, so all its environments must have the same meshes; remeshing, cutting, and tearing are outside the current scope. Robot-learning tasks that vary object geometry run one batch per object set, in separate processes or streams on the same GPU. Tasks that randomize over tens or hundreds of meshes would need many templates and compilations (about three minutes each when not cached); compiling kernels that are generic over the template's sizes, so that templates of one topology class share code, is the obvious remedy. Variation in initial state, controls, materials, and friction is unrestricted. Array indices are 32-bit, which limits a batch to roughly $10^4$ environments of the sizes we tested; the cloth-fold PPO collector, for example, cannot allocate its contact-term arrays at $B{=}4{,}096$.

\paragraph{Remaining synchronization.} Refill acts between Newton rounds; within a round, the PCG stage and the line search wait for their slowest environment, and at the end of every window environments that finish early idle while the last ones complete (\cref{fig:hetero}b). Independent Krylov streams, or preconditioners that equalize iteration counts~\citep{wang2026agipc}, could remove it. Faster per-environment solvers compose with our organization: none of our design decisions depends on how an environment's Newton direction is computed.

\paragraph{Beyond physics.} End-to-end gains are bounded by the work outside the simulator. For the visual $\pi_0$ policy, rendering and inference take 75\% of \system's time and the gain falls to 1.8\x (\cref{fig:app_stages}); batched rendering, already used in large-batch RL simulators~\citep{shacklett2021large}, and batched inference would complement batched simulation. Where physics dominates, the gain is the simulator's on that workload: in the cloth-fold PPO loop, physics takes 96\% of the time, but an action spans only three steps, so refill adds nothing, and training is 3.8\x faster in throughput and 3.2\x in time to target. Where each environment is large, a few already saturate the GPU: on LeHome's garment fold (a garment of 43,632 degrees of freedom and two arms per environment), \cgq runs fastest with one environment per process and is 1.05\x faster than \system end to end, consistent with the parity that the size sweep of \cref{sec:eval-generality} finds at this size. In our PPO collectors, refill operates within a control step: every environment contributes equally many transitions per update, from a fixed set of episodes, so refill changes wall time but not the sample distribution. An asynchronous collector that let fast environments act more often would over-represent short, often successful episodes, the bias that \cref{sec:applications} discusses for evaluation.

\paragraph{Beyond IPC\@.} We develop and evaluate decision-local batching for IPC. The same organization may be useful for other repeated-structure solvers with nested, data-dependent iterations, but its benefit is solver-dependent and remains to be demonstrated. Implicit finite elements with penalty contact share IPC's Newton--CG--line-search structure; augmented-Lagrangian trajectory optimizers, which GPU motion generators batch with shared iteration counts~\citep{sundaralingam2023curobo}, nest a Gauss--Newton loop inside multiplier updates; interior-point solvers, design optimization, and sensitivity analysis over many variants of one scene, as in differentiable IPC workloads~\citep{huang2026diffgipc}, are other plausible candidates. In such systems, $\rho$ of \cref{eq:waste} can indicate how much lockstep iteration-count variability exists, but the actual gain also depends on the cost of partially filled rounds, opportunities for sharing structure, and the memory behavior of the solver.

\section{Conclusion}
Robot learning needs many simulations that look alike and behave differently. \system turns this property into performance by sharing what rollouts have in common and keeping private what makes each of them distinct: one template of structure, laid out so that repeated topology becomes coalesced memory access, and per-environment numerical decisions that never couple one rollout to another. On manipulation workloads, \system is 4.9--10.0\x faster than a CUDA-graph IPC implementation at its best multi-process configuration and 7.3--8.2\x faster than the public libuipc library; in our tests, batch composition changes no trajectory beyond the nondeterminism of a repeated solo run, and the speedup carries through to evaluating and training robot policies. The key is not per-environment stopping alone: a conventional tiled solver given the same local convergence decisions recovers only a small fraction of the gap. The larger gain comes from making independent progress compatible with one wide, structure-sharing execution. Although we evaluate this organization only for IPC, our results suggest that the same separation---share structure, keep state-dependent decisions local---may be useful in other nested iterative solvers.

\bibliographystyle{ACM-Reference-Format}
\bibliography{references}
\appendix
\section{\texorpdfstring{\cgq}{\cgqplain} configuration search}
\label{app:search}
\cgq's linear solver is a component-partitioned PCG: at every Newton iteration it splits the Hessian into its connected components, tests each for convergence on its own, and masks converged components out of the sparse product, the preconditioner, and the vector updates. \Cref{tab:search} lists the strict throughput of every configuration of \cgq that we ran on the three benchmarks on an H100: the one-scene configuration ($P{=}1$, without MPS) and $P$ MPS processes with $K$ tiled environments each. $K$ doubles until a configuration fails, two consecutive points fall below 90\% of the curve's best so far, or $PK$ would exceed 4,096; \system's batch size follows the same rule with a cap of 16,384. The rule is applied to single screening runs, so a row of medians can extend past two cells below 90\% (cloth, $P{=}1$). Cells are medians of three interleaved timing rounds (\cref{sec:eval-setup}) for the configurations of \cref{fig:scaling} and otherwise single screening runs; ``oom'' marks configurations that run out of memory. Intermittent launch faults of \cgq under MPS (7 of 299 runs) were retried once, and all records are kept; one cloth configuration ($8{\times}16$) faulted on both attempts (``fault''). \cgq's single-environment scenes ran reliably for every benchmark; only its multi-process packings and its port of the rod tasks of \cref{sec:applications} occasionally aborted with native CUDA errors, which we report and never filter. Every configuration whose screening run came within 3\% of the best median received these three rounds; the best configurations (bold) are those of \cref{tab:benchmarks}. On the cloth press 16 processes of 8 environments tie 24 processes of 4 within run-to-run noise (1,495 and 1,490), 32 processes peak slightly lower, and 48 do not fit.

\begin{table*}[t]
\caption{Strict env-steps/s of every \cgq configuration on an H100 ($P$ processes, rows; $K$ environments per process, columns). Bold: best configuration (\cref{tab:benchmarks}).}
\label{tab:search}
\centering\fontsize{7pt}{8.4pt}\selectfont\setlength{\tabcolsep}{2.6pt}
\begin{tabular}{@{}lrrrrrrrrrrr@{}}
\toprule
\multicolumn{12}{@{}l}{\emph{Block press}} \\
\midrule
$P\backslash K$ & 1 & 2 & 4 & 8 & 16 & 32 & 64 & 128 & 256 & 512 & 1024 \\
\midrule
1 & 217 & 388 & 658 & 1,009 & 1,486 & 2,011 & 2,598 & 2,745 & 2,575 & 2,177 & 1,719 \\
2 & 358 & 631 & 975 & 1,588 & 2,301 & 2,917 & 3,356 & 3,267 & 2,732 & 2,379 &  \\
4 & 591 & 957 & 1,549 & 2,481 & 3,220 & 3,724 & 3,893 & 3,536 & 3,027 & 2,463 &  \\
8 & 798 & 1,360 & 2,259 & 3,258 & 3,890 & 4,197 & 4,079 & 3,658 & 3,027 &  &  \\
16 & 917 & 1,584 & 2,493 & 3,518 & 4,109 & \textbf{4,294} & 4,084 & 3,647 & 2,964 &  &  \\
24 & 912 & 1,554 & 2,466 & 3,528 & 4,070 & 4,252 & 4,037 & 3,615 &  &  &  \\
32 & 889 & 1,515 & 2,412 & 3,485 & 4,011 & 4,190 & 4,002 & oom &  &  &  \\
48 & oom &  &  &  &  &  &  &  &  &  &  \\
\bottomrule
\end{tabular}\\[6pt]
\begin{tabular}{@{}lrrrrrrrrr@{}}
\toprule
\multicolumn{10}{@{}l}{\emph{Cloth press}} \\
\midrule
$P\backslash K$ & 1 & 2 & 4 & 8 & 16 & 32 & 64 & 128 & 256 \\
\midrule
1 & 229 & 330 & 473 & 582 & 549 & 537 & 516 & 442 & 372 \\
2 & 292 & 553 & 748 & 734 & 760 & 685 & 589 & 491 &  \\
4 & 511 & 867 & 944 & 1,113 & 986 & 831 &  &  &  \\
8 & 756 & 1,060 & 1,260 & 1,398 & fault &  &  &  &  \\
16 & 859 & 1,237 & 1,421 & \textbf{1,495} & 1,179 & 896 &  &  &  \\
24 & 865 & 1,247 & 1,490 & 1,467 & 1,160 & 891 &  &  &  \\
32 & 840 & 1,246 & 1,453 & 1,444 & 1,146 & 892 &  &  &  \\
48 & oom &  &  &  &  &  &  &  &  \\
\bottomrule
\end{tabular}\qquad
\begin{tabular}{@{}lrrrrrrrrr@{}}
\toprule
\multicolumn{10}{@{}l}{\emph{Franka cloth fold}} \\
\midrule
$P\backslash K$ & 1 & 2 & 4 & 8 & 16 & 32 & 64 & 128 & 256 \\
\midrule
1 & 147 & 232 & 407 & 533 & 719 & 842 & 864 & 715 & 591 \\
2 & 220 & 356 & 561 & 750 & 954 & 1,034 & 1,029 & 830 & 667 \\
4 & 346 & 554 & 868 & 1,050 & 1,200 & 1,248 & 1,101 & 907 &  \\
8 & 452 & 740 & 1,039 & 1,259 & 1,308 & 1,297 & 1,122 & 909 &  \\
16 & 472 & 766 & 1,122 & 1,293 & \textbf{1,356} & 1,280 & 1,110 & 905 &  \\
24 & 459 & 724 & 1,079 & 1,252 & 1,310 & 1,269 & 1,114 & oom &  \\
32 & 436 & 696 & 1,058 & 1,289 & 1,291 & oom &  &  &  \\
48 & oom &  &  &  &  &  &  &  &  \\
\bottomrule
\end{tabular}
\end{table*}

\Cref{tab:search-libuipc} lists all 118 libuipc configurations that we ran on the block and cloth presses under the same rule: the one-scene configuration, one process advancing four worlds in host threads, and $P$ MPS processes with $K$ tiled environments each. Every row ends at a failure or after two consecutive points below 90\% of its best. More than 16 processes do not fit, because each libuipc instance holds about 4.6\,GB of GPU memory even for one environment. Worlds in one process do not run concurrently: with libuipc's default CUDA-graph PCG the threads abort, and with graph capture off (the row shown) four worlds of $K$ environments are no faster than one scene of $4K$. The five best configurations received three interleaved pairs; the best ones (bold) are those of \cref{tab:benchmarks}.

\begin{table*}[t]
\caption{Strict env-steps/s of every libuipc configuration on an H100 at the tolerances of \cref{tab:parameters} ($P$ MPS processes, rows; $K$ environments tiled in one libuipc scene per process, columns; ``1 (4 worlds)'': one process advancing four worlds of $K$ environments each in host threads). Bold: best configuration (\cref{tab:benchmarks}); $^\dagger$: the other timed candidates; bold and $^\dagger$ cells are medians of three interleaved pairs, all others single screening runs. oom: GPU memory exhausted (each libuipc instance holds about 4.6\,GB); nan: libuipc aborted on a diverged PCG ($\mathbf r^{\mathsf T}\mathbf z$ = NaN); segv: segmentation fault.}
\label{tab:search-libuipc}
\centering\fontsize{7pt}{8.4pt}\selectfont\setlength{\tabcolsep}{2.6pt}
\begin{tabular}{@{}lrrrrrrrrrrr@{}}
\toprule
\multicolumn{12}{@{}l}{\emph{Block press}} \\
\midrule
$P\backslash K$ & 1 & 2 & 4 & 8 & 16 & 32 & 64 & 128 & 256 & 512 & 1024 \\
\midrule
1 & 137 & 243 & 373 & 523 & 775 & 1,156 & 1,622 & 1,908 & 1,777 & 1,573 & 1,210 \\
1 (4 worlds) & 129 & 201 & 310 & 484 & 754 & 1,193 & 1,636 & 1,845 & 1,818 & 1,578 & 1,209 \\
2 & 245 & 393 & 580 & 858 & 1,301 & 1,849 & 2,352 & 2,476 & 2,111 & 1,715 &  \\
4 & 404 & 609 & 932 & 1,466 & 2,077 & 2,743 & 3,106 & 2,886 & 2,348 & 1,801 &  \\
8 & 625 & 1,015 & 1,598 & 2,341 & 3,137 & 3,751 & 3,665 & 3,101 & 2,449 &  &  \\
16 & 934 & 1,541 & 2,471 & 3,429 & 4,157\rlap{$^\dagger$} & \textbf{4,369} & nan &  &  &  &  \\
24 & oom &  &  &  &  &  &  &  &  &  &  \\
32 & oom &  &  &  &  &  &  &  &  &  &  \\
48 & oom &  &  &  &  &  &  &  &  &  &  \\
\bottomrule
\end{tabular}
\\[6pt]
\begin{tabular}{@{}lrrrrrrrrrrr@{}}
\toprule
\multicolumn{12}{@{}l}{\emph{Cloth press}} \\
\midrule
$P\backslash K$ & 1 & 2 & 4 & 8 & 16 & 32 & 64 & 128 & 256 & 512 & 1024 \\
\midrule
1 & 135 & 242 & 385 & 613 & 735 & 907 & 1,018 & 1,025 & 1,024 & 899 & 839 \\
1 (4 worlds) & 130 & 224 & 331 & 530 & 747 & 946 & 1,044 & 1,039 & 1,022 & 919 & segv \\
2 & 246 & 403 & 649 & 856 & 1,117 & 1,222 & 1,246 & 1,201 & 1,054 & 953 &  \\
4 & 418 & 700 & 897 & 1,292 & 1,489 & 1,495 & 1,359 & 1,192 & 1,077 &  &  \\
8 & 723 & 1,024 & 1,374 & 1,736 & 1,799 & 1,613 & 1,362 &  &  &  &  \\
16 & 983 & 1,501 & 1,823\rlap{$^\dagger$} & \textbf{2,056} & 1,902\rlap{$^\dagger$} & oom &  &  &  &  &  \\
24 & oom &  &  &  &  &  &  &  &  &  &  \\
32 & oom &  &  &  &  &  &  &  &  &  &  \\
48 & oom &  &  &  &  &  &  &  &  &  &  \\
\bottomrule
\end{tabular}
\end{table*}

\section{Parameters of the Newton comparison}
\label{app:vbd}
Both backends read the same meshes (block: 405 vertices and 1,536 tetrahedra; cloth: $31{\times}31$ vertices, 0.3\,m), the same tool path, and SI units. The block uses VBD's stable Neo-Hookean model with \system's Lam\'e parameters ($E{=}10^5$\,Pa, $\nu{=}0.4$, 1,000\,kg/m$^3$) and no damping. The cloth uses the same areal density (0.4\,kg/m$^2$) and a dihedral bending stiffness fitted to \system's quadratic bending energy~\citep{bergou2006quadratic}; its membrane stiffness (7,000\,N/m) is calibrated to match the maximum stretch of \system's strain-limited Baraff--Witkin membrane, because a small-strain energy match let VBD's cloth stretch 3.7-fold under the press. The tool is the same stable Neo-Hookean box on both backends, tied to its path by springs of \system's position-constraint stiffness. Friction uses $\mu{=}0.5$ and $\varepsilon_f{=}10^{-2}$\,m/s with the same $C^1$ mollifier as IPC. VBD's contact distance equals \system's $\hat d{=}2$\,mm, and its nominal contact stiffness $10^5$\,N/m is the value of \system's initial barrier stiffness; the floor uses VBD's log-barrier option. Apart from the cloth membrane, no parameter was tuned for agreement. VBD's default on the presses is ten substeps of ten iterations at this contact stiffness, the substeps and iterations of Newton's cloth-bending example (its cloth examples use 10--20 substeps of 4--10 iterations); its curve varies one to twenty substeps, 2 to 50 iterations, and contact stiffnesses of $10^4$, $3{\cdot}10^4$, $10^5$, $3{\cdot}10^5$, and $10^6$\,N/m (125 configurations per press). \system uses the parameters of \cref{tab:parameters} and varies the Newton tolerance from $10^{-3}$ to $3{\cdot}10^{-1}$\,m/s at the time step $h$; two of these tolerances also run at $h/2$.

\paragraph{Version.} All Newton runs use commit \texttt{f6fc82a8} of the development fork \href{https://github.com/dylanturpin/newton-collab}{\nolinkurl{github.com/dylanturpin/newton-collab}}, which contains the STEEL rigid-body solver and the cloth and block example. Deformables run on VBD and robots on STEEL, coupled through Newton's proxy solver with one coupling iteration per substep.

\paragraph{Cable and cloth-block scenes.} These are two examples of that fork: an FR3 arm picks a 0.38\,m cable of 19 segments and places it at a target, and an FR3 arm lifts a 0.25\,kg block out of a cloth hammock of $20{\times}20$ cells, pinned along two edges, and places it on a table. We changed six things on Newton's side: the cable's twist stiffness and twist damping are zero, as in \system's rod; every environment draws its own inputs (initial offset and yaw of the cable or block, place target, and keyframe durations); the examples' inverse kinematics, which never reads the simulated state, is precomputed into a table of joint targets that both systems share; the cloth-block example's correction of the grasp target from the settled block is computed for each environment, not copied from the first one, and its time is excluded on both systems; the number of STEEL iterations is configurable for the converged reference; and one batch is timed. Success is each example's own test: the cable's midpoint within 10\,mm of the target; the block within 20\,mm of the target and 5\,mm of the table height, with the cloth rebounding by more than 10\,mm. \system models the arm as affine bodies with the same masses, collision shapes, and drive gains; it omits Newton's effort limits, which its controller would apply differently and which never bind in these tasks. The cable is a discrete elastic rod with the capsule chain's mass and its stiffness per segment ($EA{=}2.0$\,N, $EI{=}8{\cdot}10^{-6}$\,N\,m$^2$, no twist stiffness); the hammock is a shell with the small-strain moduli of Newton's membrane ($Et{=}2{,}667$\,N/m, $\nu{=}1/3$) and a fitted bending stiffness; neither has Newton's material damping. Contact uses the Franka values of \cref{tab:parameters} with Newton's friction coefficient of 1. Newton's defaults are the examples' ten substeps with 20 (cable) or 10 (cloth-block) VBD iterations and 24 STEEL iterations; its curves vary 2 to 40 substeps and 5 to 50 VBD iterations, and include three configurations with four times stiffer contact (23 configurations each). \system's default is $\varepsilon_v{=}0.05$\,m/s at one step per 60\,Hz frame.

\paragraph{Cable configurations and timing.} Nine of Newton's 23 cable configurations succeed in fewer than 90\% of the environments: all four with two substeps, all four with five substeps at the default contact stiffness, and twenty substeps of five iterations. The cheapest of the other 14 is the matched configuration, five substeps of 20 iterations with four times stiffer contact. \system's default satisfies the strict contract in all paired runs of both cable ratios (7.0--7.1\x and 3.6\x in each of the three pairs). Three of its other curve points do not: at $\varepsilon_v{\le}0.003$\,m/s, and at $h/2$ with $\varepsilon_v{=}0.01$\,m/s, some steps of the timed run exhaust the Newton, line-search, or PCG budget; they are committed and counted, as in \cref{sec:applications}.

\paragraph{Fold.} The fold uses the Panda arm and the cloth of \cref{tab:benchmarks} on both systems; in Newton the arm runs on STEEL. With the original scene's controls, authored under IPC dynamics, Newton completed no fold, so we redesigned the scene on 16 development environments before the comparison: the riser that supports the flap is narrowed (footprint $[0.495, 0.545]{\times}[-0.025, 0.015]$\,m), the gripper squeezes with 3\,N, and each environment corrects its grasp target from the position of the settled flap vertex. A fold succeeds if the cloth's top rises above 0.75\,m during the fold and its largest $x$ coordinate ends below 0.50\,m; the fold error is the RMS horizontal distance of three folded-edge vertices from where the commanded transport of the pinch would put them. Newton's default, as on the presses, is ten substeps of ten iterations (20 configurations). Loosened to $\varepsilon_v{\ge}0.1$\,m/s, \system's pinch slips and at most 14\% of the folds succeed.

\paragraph{References and convergence.} The IPC reference is \system at $h/4$ with $\varepsilon_v{=}10^{-3}$\,m/s; it is also \system's own reference solution on the presses. Halving \system's time step from $h/2$ to $h/4$, from $h/4$ to $h/8$, and from $h/8$ to $h/16$ changes its result by 0.61, 0.45, and 1.29\,mm on the block press and by 0.19, 0.13, and 6.21\,mm on the cloth press (90th percentile over the 128 environments), and $\varepsilon_v{=}10^{-4}$\,m/s at $h/4$ changes it by 0.02\,mm. NVIDIA Newton's own reference solutions are four substeps of 200 iterations at each contact stiffness on the presses, and 40 substeps of 100 iterations with 96 STEEL iterations on the cloth-block scene and the fold. On the block press, doubling VBD's iterations from 100 to 200, 400, and 800 changes its result by 0.24, 0.02, and 0.005\,mm at $10^4$\,N/m, by 0.24, 0.14, and 0.11\,mm at $3{\cdot}10^4$\,N/m, by 0.29, 0.22, and 0.17\,mm at $10^5$\,N/m, by 0.98, 0.52, and 0.43\,mm at $3{\cdot}10^5$\,N/m, and by 3.7, 1.9, and 1.5\,mm at $10^6$\,N/m; beyond 200 iterations, it changes the errors of the block-press configurations quoted in \cref{sec:eval-cross} by at most 0.04\,mm. On the cloth press, each doubling changes VBD's result by 2.6--5.0\,mm at every stiffness. VBD does not converge under substep refinement on the cloth press: twenty substeps of 100 iterations differ from its reference by 42\,mm. In the controlled sweep of \cref{sec:eval-generality}, timed on an H100, VBD's grid has 50 configurations per cell; at $128^2$ it was cut off by run time, and a configuration that finished later was slightly more accurate (14.1 and 7.4\,mm, against 15.3 and 7.5\,mm) and was not timed. VBD's own reference solution in each cell is four substeps of 200 iterations at each contact stiffness.

\begin{figure}[t]
\centering
\includegraphics[width=\columnwidth]{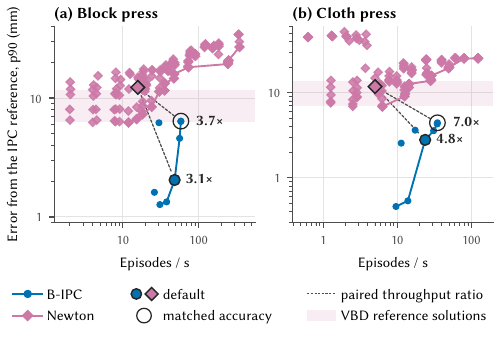}
\caption{\textbf{\system stays closer to the IPC reference across the press sweeps.} The presses of \cref{fig:newton}a,b against the IPC reference. The band marks the distance of VBD's reference solutions (four substeps of 200 iterations, one per contact stiffness) from the IPC reference; rings mark \system's configurations matched to VBD's default. Markers and lines as in \cref{fig:newton}.}
\Description{Two log-log scatter plots of the error from the IPC reference against episodes per second for the block and cloth presses. B-IPC's points lie below Newton's cloud, Newton's points lie above about six millimeters, and a shaded band between about six and fourteen millimeters marks VBD's reference solutions.}
\label{fig:newton-ipc}
\end{figure}

\paragraph{The presses against the IPC reference\@.} \Cref{fig:newton-ipc} shows the press configurations of \cref{fig:newton} against the IPC reference. VBD's reference solutions lie 6.3--11.6\,mm (block) and 6.9--13.8\,mm (cloth) from it, closest at the lowest stiffness. On this axis, VBD's defaults are at 12.3 and 11.8\,mm and its most accurate configurations at 6.2 and 6.9\,mm, against 2.0 and 2.8\,mm for \system's defaults. Loosened to $\varepsilon_v{=}10^{-1}$ and $3{\cdot}10^{-1}$\,m/s, \system remains closer to the IPC reference than VBD's default (6.4 and 4.4\,mm) and is 3.7\x and 7.0\x faster. VBD overtakes \system's fastest configuration at errors between 18 and 20\,mm: at $3{\cdot}10^4$\,N/m, one substep of five iterations completes 245 block-press episodes/s at 19\,mm, and one of ten iterations 50 cloth-press episodes/s at 20\,mm.

\section{Benchmark scenes and controls}
\label{app:controls}
\Cref{tab:controls} gives the scenes of \cref{tab:benchmarks} and the per-environment controls. Environment $e$ draws every random quantity from its own counter-based stream keyed by $e$, so it sees the same inputs at every batch size and in both systems. Every sixteenth environment of each press is a hard case with the largest press depth and slide. All presses and the cloth drop use the parameters of \cref{tab:parameters}; the Franka fold uses those given in its caption.

\begin{table*}[t]
\caption{Benchmark scenes and per-environment controls. $U[a,b]$: uniform; $U\{0,\ldots,n\}$: uniform integer.}
\label{tab:controls}
\centering\footnotesize\setlength{\tabcolsep}{3pt}
\begin{tabular}{@{}p{0.11\textwidth}p{0.42\textwidth}p{0.44\textwidth}@{}}
\toprule
Scene & Geometry and material & Per-environment controls \\
\midrule
Block press & $0.10{\times}0.10{\times}0.05$\,m block, $8{\times}8{\times}4$ cells (405 vertices, 1,536 tetrahedra), stable Neo-Hookean, $E{=}10^5$\,Pa, $\nu{=}0.4$, 1,000\,kg/m$^3$; pusher: $0.04{\times}0.04{\times}0.02$\,m box, $E{=}10^6$\,Pa, driven by position springs; 360 steps of 1/60\,s & contact point offset $U[-15,15]$\,mm in $x$ and $y$; press depth $U[8,15]$\,mm; slide $U[20,40]$\,mm in direction $U[0,2\pi)$; time shift $U\{0,\ldots,20\}$ steps; the pusher touches at step $60{+}j$, presses until $120{+}j$, slides until $240{+}j$, and lifts until $300{+}j$; hard cases: 20\,mm depth, 50\,mm slide \\
Cloth press & $0.30{\times}0.30$\,m cloth, $30{\times}30$ cells (961 vertices, 1,800 triangles), Baraff--Witkin membrane $E{=}10^5$\,Pa with strain limiting, quadratic bending, 1\,mm one-sided thickness (a 2\,mm shell), 200\,kg/m$^3$ (0.4\,kg/m$^2$); same pusher; 360 steps of 1/60\,s & contact point offset $U[-50,50]$\,mm; press depth $U[2,4]$\,mm; slide and timing as for the block press; hard cases: 6\,mm depth, 50\,mm slide \\
Franka cloth fold & Franka Panda with 7 revolute and 2 prismatic joints and PD drives; arm links, table, and riser as affine bodies; $0.24{\times}0.24$\,m cloth, $24{\times}24$ cells (625 vertices), Baraff--Witkin $E{=}2{\cdot}10^4$\,Pa, hinge bending~\citep{grinspun2003shells}; 1,130 steps of 0.01\,s & one scripted fold (joint targets and feed-forward forces at 13 keyframes) applied with a start delay $U\{0,\ldots,10\}$ steps \\
Cloth drop & $0.30{\times}0.30$\,m cloth, $65{\times}65$ vertices (8,192 triangles), Baraff--Witkin $E{=}10^5$\,Pa, hanging vertically with its lower edge 0.10\,m above the floor; 180 steps of 1/60\,s & translation $U[-10,10]$\,mm in $x$ and $y$, height $U[-5,5]$\,mm; initial out-of-plane bulge velocity of amplitude $U[0.75,1.25]$\,mm/s with random sign; all environments touch the floor near step 9 \\
\bottomrule
\end{tabular}
\end{table*}

\begin{table}[t]
\caption{Solver settings in the \taccel comparison (\system{}\,/\,\taccel). Where one value is given, both systems use it.}
\label{tab:taccel}
\centering\footnotesize\setlength{\tabcolsep}{2pt}
\begin{tabular}{@{}p{0.25\columnwidth}p{0.355\columnwidth}p{0.355\columnwidth}@{}}
\toprule
Setting & Peg insertion & Block press \\
\midrule
Batch size & 8,192 / 4,096 & 8,192 / 4,096 \\
Time step; steps & 0.02\,s; 200 & 1/60\,s; 360 \\
$\hat d$; $\kappa$; $\mu$ & 0.1\,mm; $10^5$\,N/m; 0.2--1.0 & 1\,mm; $10^5$\,N/m; 0.4 \\
Newton tolerance & $10^{-2}$\,m/s & $10^{-2}$\,m/s \\
Newton budget & 1,024, step fails / 40, accepted & 1,024, step fails / 20, accepted \\
Linear solver & MAS-PCG / block-Jacobi CG & MAS-PCG / block-Jacobi CG \\
Linear tolerance & $10^{-4}$ / $10^{-5}$ (relative) & $10^{-4}$ / $10^{-5}$ (relative) \\
Linear budget & 1,024 / 500 & 1,024 / 500 \\
Precision & FP64 (FP32 MAS) / FP64 & FP64 (FP32 MAS) / FP64 \\
Contact storage & presets / $2^{23}$ pairs, grows & same \\
\bottomrule
\end{tabular}
\end{table}

\section{Solver settings of the \texorpdfstring{\taccel}{Taccel} comparison}
\label{app:taccel}
\Cref{tab:taccel} lists the settings behind \cref{sec:eval-cross}. Both systems run the same episode pool, time step, material, contact parameters, and Newton tolerance. They differ in the linear solver and in how budgets are enforced: \system fails a step that exhausts a budget (C2), whereas \taccel accepts steps that reach its Newton cap, which favors \taccel. \taccel's Newton cap is its released default, 20, on the block press and our choice, 40, on the peg insertion, whose steps need up to 24 iterations; its CG budget is the 500 iterations that its released core applies, on both tasks. With a CG budget of 40 passed to the core by a wrapper, which its solves on the block press reach, \taccel completes 1,042 instead of 485 env-steps/s there (\system: 19.3\x). Parts of \taccel's Newton loop (its convergence norm, line-search rule, and stiffness adaptation) are in a closed core and could not be inspected.

\section{Throughput accounting details}
\label{app:throughput}
This appendix supports \cref{sec:eval-setup,sec:eval-throughput}.

\paragraph{Timing window.} For a multi-process configuration, the window runs from the first process's start to the last process's end after a common start barrier, and the start skew (median 3\,ms) is charged to that configuration.

\paragraph{Iteration accounting.} The wall time per environment-PCG iteration is the total wall time of all stages divided by the number of environment-PCG iterations. For both systems we count the PCG iterations that each environment performs: \cgq's partitioned PCG ends the solve of each environment of a tiled scene on its own, so the scene's loop count, which follows its slowest component, would overstate its work. A \cgq scene step that reaches a Newton, PCG, or line-search limit counts none of its environments.

\paragraph{Matched packing and per-stage profiles.} Run as \cgq's best packings ($16{\times}32$, $16{\times}8$, and $16{\times}16$ MPS processes), \system needs the same iterations (within 0.5\%), but each environment-PCG iteration costs 4.4\x, 6.5\x, and 4.1\x more; at that matched packing, an iteration costs 1.2\x less than \cgq's on the block press and 1.6\x and 1.4\x more on the cloth press and the Franka fold, and \system is 1.90\x, 1.54\x, and 1.18\x faster. Profiles of both systems at their best configurations locate the per-iteration advantage in the preconditioner apply (2.5--4.3\x cheaper per environment-PCG iteration), the dot products and vector updates (22--38\x: five fused kernels per iteration for the whole batch instead of about nine in every process), and the work per Newton iteration (4.0--5.9\x), not in the sparse operator (0.36--1.18\x). The stages hold different work: \cgq's operator stage is one sparse product with an assembled matrix that includes contact, while \system's applies the block rows and the contact terms and computes $p^\top A p$ in the same kernel, and on the Franka fold also the affine bodies' world-vertex products.

\paragraph{Matched error.} Against a reference converged to $\varepsilon_v/100$ (\cgq with one environment per scene; \system at $\varepsilon_v/100$ agrees with it within 0.7\,\textmu{}m in every statistic), the median final-state RMS error over 192 environments of each press is 146\,\textmu{}m and 1.1\,mm for \system at $\varepsilon_v$ on the block and cloth presses, 145\,\textmu{}m and 1.1\,mm for \cgq with one environment per scene, and 39\,\textmu{}m and 0.22\,mm for \cgq's best packings, whose scene-wide test keeps converged environments iterating. \system reaches 24 and 53\,\textmu{}m at $\varepsilon_v/5$, the largest of the tolerances $\varepsilon_v$, $\varepsilon_v/2$, $\varepsilon_v/5$, and $\varepsilon_v/10$ at which its error does not exceed that of the packings. At $\varepsilon_v/5$ it completes 25,215 and 7,347\,\envsteps, 5.9\x (range 5.86--5.87 over three rounds) and 5.0\x (4.88--5.10) the rate of \cgq's best packing at $\varepsilon_v$, with 1.2\x fewer Newton iterations per environment step on both presses. With both systems at $\varepsilon_v/10$ (errors of 10 and 14\,\textmu{}m against 1.8 and 2.7\,\textmu{}m), \system is 10.2\x and 10.8\x faster than the fastest \cgq packing at that tolerance ($16{\times}32$ and $24{\times}4$).

\begingroup\emergencystretch=1em %
\paragraph{Cost against batch width.} One \cgq scene is also a wide batch in one process, so we compare the two systems at equal width, per environment-Newton iteration (the one-scene runs did not record per-environment PCG counts). \cgq's wall time per environment-Newton iteration falls with the number $K$ of tiled environments and levels off: on the block press, 140\,\textmu{}s at $K{=}32$, 78\,\textmu{}s at 128, and 63.6\,\textmu{}s at 1,024, against 67\,\textmu{}s in the $16{\times}32$ packing; on the cloth press, 395\,\textmu{}s at $K{=}8$ and 114\,\textmu{}s at 256, against 148\,\textmu{}s at $16{\times}8$; on the Franka fold, 189\,\textmu{}s at $K{=}64$ and 176\,\textmu{}s at 256, against 174\,\textmu{}s at $16{\times}16$. The last doubling of $K$ lowers it by 4\%, 10\%, and 1\%. \system in one process costs 106, 45, and 15.4\,\textmu{}s at $B{=}32$, 128, and 1,024 on the block press and 12.5\,\textmu{}s at $B^\star$; on the cloth press, 46\,\textmu{}s at $B{=}256$ and 28\,\textmu{}s at $B^\star$; on the Franka fold, 100\,\textmu{}s and 59\,\textmu{}s. At equal width, \cgq's iteration costs 1.31\x, 1.06\x, and 0.70\x as much as \system's at the per-process widths of its packings ($K{=}32$, 8, and 16), and 4.14\x, 2.46\x, and 1.77\x as much at the widest scenes we ran (1,024, 256, and 256); widening \system from there to $B^\star$ lowers its cost by another 1.23\x, 1.65\x, and 1.68\x. \cgq's throughput peaks at narrower scenes ($K{=}128$, 8, and 64), because the scene-wide test raises its Newton iterations per step with $K$. The equal-width ratios compare runs of two sessions, between which \system's cost at $B^\star$ differs by at most 0.4\%.\par
\endgroup

\paragraph{Placement of the hard environments.} Placing the hard environments of the presses at random instead of every sixteenth, removing them, or sorting them into one tiled \cgq scene changes these speedups by at most 3.6\%.

\section{Details of the libuipc comparison}
\label{app:libuipc}
\paragraph{Implementation differences.} libuipc replaces the requested barrier stiffness by its own scene-dependent value, uses a projected full barrier Hessian, and discretizes cloth bending differently. The observed ranges of the paired ratios at the best configurations are 8.21--8.28\x and 7.27--7.31\x.

\paragraph{Single-environment rates.} At $B{=}1$, libuipc completes 137 and 135 env-steps/s, 0.63\x and 0.59\x \cgq's 217 and 229 (\system: 284 and 285): it takes fewer Newton iterations per step than \cgq but drives them from the host, so each costs about twice as much; batching closes this gap.

\paragraph{Final states.} libuipc's final states differ from \system's by a median of 0.28 and 0.88\,mm, against 0.13 and 1.2\,mm between \cgq runs at $\varepsilon_v$ and $\varepsilon_v/10$ on the 256 block-press and 128 cloth-press environments of this comparison.

\paragraph{Looser defaults.} With both systems at libuipc's own looser defaults, \system remains 3.7\x and 6.0\x faster (\cref{sec:eval-cross}). The block-press ratio falls because at this tolerance libuipc stops after one Newton iteration on most steps, whereas \system, like \cgq, always takes at least two, and libuipc's semi-implicit early exit has no counterpart in \system; at these settings libuipc's final states differ from \system's converged ones by 1.3 and 3.1\,mm.

\section{\texorpdfstring{\taccel}{Taccel} comparison: batch dependence and solver work}
\label{app:taccel-audit}
\paragraph{H100.} On an H100, where \taccel also runs, the ratios are 57.2\x and 51.4\x: \taccel's step speeds up only 1.6\x from the RTX PRO 6000, less than the memory-bandwidth ratio, so it is bound by memory traffic, launches, and host work rather than by double-precision throughput.

\paragraph{Batch dependence.} On the block press, the benchmark's original gate, which requires the two systems' block trajectories to agree within 5\,mm, fails by up to 21\,mm on the 21 audited episodes (the final states of the 4,097 episodes that both timing runs store differ by up to 26\,mm), and single-environment runs show why. At $B{=}1$ the two simulators agree within 5\,mm on all 21 audited episodes. On these episodes, two single-environment runs of \taccel differ by at most 0.36\,mm (median below 0.001\,mm), and the final states of two batched runs of the same pool by at most 0.6\,mm, but its batched trajectories differ from its single-environment ones by 8.5--21\,mm on every episode it ran in a batch; the difference depends on the batch size (two episodes: 0.000, 6.8, 5.7, and 14.8\,mm, and 6.2, 7.1, 8.8, and 20.9\,mm, at $B{=}2$, 16, 256, and 4,096) and persists, at 10--21\,mm, with its collision buffer fixed at $2^{25}$ pairs from the start; \system's batched and single-environment trajectories agree within 0.05\,mm (maximum deviation over the trajectory with this material).

\paragraph{Solver work.} The two systems also differ in solver work: on a contact-heavy segment, \taccel's block-Jacobi CG takes 330 and 93 iterations per Newton iteration on the peg insertion and the block press, against 14 and 26 for \system's MAS-preconditioned PCG, and costs 0.55--0.96\,\textmu{}s per environment and iteration against 0.23\nobreakdash--0.30\,\textmu{}s, but preconditioning does not explain the ratios: with \taccel's block-Jacobi preconditioner, which on these small scenes is faster than MAS, \system is 66\x and 44\x faster, and also with \taccel's CG stopping rule and budget, about 39\x and 25\x. The remaining gap comes from Newton behavior (\taccel needs about 3\x more Newton iterations per step), batching, and implementation.

\paragraph{Gates, episode pool, and contact storage.} The peg insertion's original response gates require the final peg position within 1\,mm and the pad deformation within 0.5\,mm of each other. The pool has $4{\times}8{,}192{+}1$ episodes, so that both systems also run a real tail batch. In all runs of this comparison, \system uses fixed per-environment contact presets (\cref{tab:parameters}), which no environment exceeded; \taccel preallocates a batch-wide collision buffer of $2^{23}$ pairs and doubles it on demand (to $2^{25}$ at $B{=}4{,}096$), and we used its default.

\section{Batch-invariance and contract tests}
\label{app:invariance}
This appendix gives the tests behind \cref{sec:eval-quality}. Where we call a rule or margin fixed in advance, it was committed to our repository before the runs it judges.

\paragraph{Final-state tails.} Over the 192 probes of each task, the final-state RMS difference between a batched run and the probe's first solo run has 90th percentiles of 2.2 and 1.4\,\textmu{}m ($B{=}64$ and $4{,}096$) on the block press, 158 and 172\,\textmu{}m on the cloth press, and 18.8 and 22.4\,mm on the Franka fold, against 1.4\,\textmu{}m, 207\,\textmu{}m, and 19.7\,mm between two solo runs; 99th percentiles of 89 and 89\,\textmu{}m, 598 and 596\,\textmu{}m, and 75.3 and 74.2\,mm, against 29\,\textmu{}m, 662\,\textmu{}m, and 76.7\,mm; and maxima of 112 and 230\,\textmu{}m, 810 and 867\,\textmu{}m, and 78.5 and 80.8\,mm, against 102\,\textmu{}m, 966\,\textmu{}m, and 79.2\,mm. Neither a two-sided Kolmogorov--Smirnov test nor a one-sided Mann--Whitney test detects a difference between the batched and the solo distributions on any task ($p\ge0.60$ and $p\ge0.30$). The block-press tail comes from four probes that each end in one of two final states 88--230\,\textmu{}m apart; one of them differs by 102\,\textmu{}m between its two solo runs. No capacity event occurred in any probe run.

\paragraph{Task outcomes.} For the pick-cube, peg-insertion, and cable-routing episodes of \cref{sec:eval-quality}, the flip rate between a solo and a batched run minus that between two solo runs is $-0.6$, $+0.3$, and $-0.2$ percentage points, with 90\% confidence intervals of $[-1.1,-0.1]$, $[-0.8,1.5]$, and $[-0.5,0.1]$ (normal approximation; bootstrap intervals agree), so they are equivalent to zero within the margin of $\pm2$ points fixed before the runs (two one-sided tests per task, $p\le0.008$; $p\le0.023$ with a Bonferroni correction for the three tasks), and no McNemar test detects a direction. The pick-cube interval excludes zero only without that correction ($[-1.3,0.1]$ with it), and in the direction opposite to a coupling: the batched run agrees with the solo runs slightly more often than they agree with each other. On the cable routing, where two solo runs flip 0.4\% of the episodes, the margin alone would admit a sixfold flip rate, but the interval of the difference excludes it. Across nine batched runs of the same 2,048 episodes, the flip rate between any two runs, solo or batched, lies between 3.0\% and 4.3\%.

\paragraph{Poison and replicate tests.} In a \emph{poison test}, 8 probe environments per task share a batch of 64 or 4,096 with neighbors that start in extreme states (5\,m/s into the ground, 3\,cm deeper presses, doubled joint targets), are re-randomized and deactivated before every Newton round, or hold NaN and Inf values. No probe becomes non-finite, while every NaN neighbor fails its step. On the presses, no discrete decision of any probe changes over 20 steps (41--42 Newton iterations)---Newton and PCG counts, active sets, line-search and CCD outcomes are identical---and the iterates differ from those of an unperturbed batch by at most 1.5\,nm, against 0.9\,nm between two unperturbed runs. On the Franka fold, the first decision flip, a contact-pair count at the first contact-rich step, occurs at the same Newton iteration with and without poison and between two solo runs, and the iterates before it differ by up to 0.07\,mm in poisoned runs and 0.03\,mm between solo runs. A replicate test whose rule was fixed before the runs (9 control and 9 poisoned runs per task) finds no neighbor effect on the block press and the Franka fold ($p\ge0.13$); on the cloth press it flags a difference ($p{=}0.03$) that comes from mixing solo and batched controls: over these steps, repeated solo runs differ from the first by less than $10^{-14}$\,m in the median, and batches of 64 differ from it by up to $1.2{\cdot}10^{-11}$\,m with or without poison (post hoc, poisoned versus unperturbed batches: $p{=}0.41$). A second rule fixed in advance, a 10\x envelope around two control runs, fires at isolated iterations where the two controls happened to round identically; it cannot bound a heavy-tailed rounding spread. The first Newton iterates and PCG residual sequences at $B\in\{1,64,4{,}096\}$ likewise match the solo run as closely as a second solo run does.

\paragraph{Deterministic mode.} On the cloth press, in the deterministic mode of \cref{sec:determinism}, we compare each step's full state (a hash of all positions and velocities) and its Newton, PCG, and line-search counts. Each of the 192 probes equals its solo run at all 360 steps in refill batches of 64, 512, 2,048, and 4,096, in a repeated batch of 4,096, and in a lockstep batch of 512 at other slots with other neighbors; so do 48 probes next to a neighbor pressed 3\,cm deeper, 48 next to a NaN neighbor, and 32 repeated solo runs (460,800 environment steps in total). Two runs of the same 4,096-environment batch are equal in all 1,474,560 environment steps. The capacity coupling of \cref{sec:determinism} occurred once: in the batches of 4,096, one neighbor overflowed a candidate buffer and its step was replayed after the buffer grew. The probes stayed bitwise equal; the proposition covers such replayed steps and excludes only steps kept after the retry budget. The mode's trajectories differ from the default build's by as much as two runs of the default build differ from each other (per-probe maximum vertex distance: median 0.25--0.26\,mm against 0.25\,mm, 90th percentile 1.6--1.9\,mm against 1.5\,mm), and every step of every unpoisoned environment satisfies C2. At $B^\star{=}2{,}048$ on an H100, in three paired rounds, the mode reaches 0.571\x the throughput of the default build (range 0.571--0.573): strict compilation accounts for 0.89\x, batch-size-independent dispatch for 0.84\x, and the fixed orderings for 0.76\x.

\paragraph{Strict completion and reference equivalence.} In the quality runs of the Franka fold (C2), one \system step also exhausts the Newton budget. Of the 120 environments in which \system is not closer to \cgq than \cgq at $\varepsilon_v$ is to \cgq at $\varepsilon_v/10$ (C3), 119 are Franka-fold environments, where the two differences are of the same size; the remaining one, a cloth-press environment with one of the longest Newton trajectories, ties at 0.64 versus 0.63\,mm. The smallest contact distance is 0.105$\hat d$ over all 8,192 environments of the \system batches. Neither solver prevents element inversion, because the stable Neo-Hookean material, unlike a barrier-type elastic energy, admits inverted elements: the same four of the first 256 block-press environments (1.6\%) invert tetrahedra in both systems, for 45 steps each. They are among the hardest (20\,mm press), and their minimum volume ratio is $-0.21$; in \system's batch of 4,096, 83 of the 256 hard environments invert in 1,429 environment steps (minimum volume ratio $-0.28$).

\section{Attribution details}
\label{app:attribution}
This appendix supports \cref{sec:eval-attribution}.

\paragraph{Capacity modes.} The three benchmark timings use preset per-task capacities for \system and growing pooled buffers for \cgq because those are the faster modes for the respective systems. With both systems growing from small capacities, the \system speedups are 8.3\x, 7.1\x, and 3.9\x on the block press, cloth press, and Franka fold; with both systems on presets they are 8.4\x, 10.1\x, and 4.9\x (9.1\x, 15.3\x, and 7.3\x when \cgq's padding is also fixed). At $B{=}4{,}096$, the Franka fold's preset contact-term buffer exceeds the array library's 32-bit index range, so the scaling point uses grown capacities and reaches 5,264--5,389 env-steps/s in two runs.

\paragraph{Refill round accounting.} At $B^\star$, the per-step round ratio $\rho$ of \cref{eq:waste} is 4.1, 8.4, and 3.1 on the block press, cloth press, and Franka fold. Over a fixed window, refill is also affected by the final drain: $\sum_k\max_e N_{e,k}/\max_e\sum_k N_{e,k}$ is 3.2, 5.4, and 3.0, and the drain accounts for about 6\%, 7\%, and 1\% of window time. Most lockstep tail rounds are cheap---in 72--85\% of them fewer than an eighth of environments remain active, and such rounds cost 1--5\% of a full round---so the throughput gain is much smaller than the raw round ratio. Weighting rounds by their measured cost predicts 1.14\x and 1.35\x of the 1.22\x and 1.72\x leave-one-out refill gains on the two presses; we have not attributed the remainder.

\paragraph{Scene-wide Newton termination.} \cgq's one-scene configuration shows the same growth of the Newton iteration count with the number of environments per scene, from 2.2--2.6 iterations per step for one environment to 9.1 for 1,024 block presses and 17.8 for 128 cloth presses. In the ablation with a scene-wide test, 0.35--4.7\% of the Newton steps that converged environments keep taking overshoot the tolerance and fail the test again, and PCG work rises from 33 to 248, 24 to 5,520, and 69 to 1,130 iterations per environment step (\cref{fig:ablation}c).

\begingroup\emergencystretch=1em %
\paragraph{Per-environment decisions in \cgq.} \looseness=-1 %
Our modification of \cgq (\cref{sec:eval-throughput}) labels the environments of a tiled scene and takes per environment the decisions that \cgq takes once per scene: the Newton stopping test and exit, the line search (each environment has its own step size and energy test; its energy is summed from the solver's per-stencil energies, and the sum over the environments equals the scene energy to rounding at every evaluation), the tolerance floor of the partitioned PCG, and the growth of the adaptive barrier stiffness. A converged environment is masked from the PCG. The patch has about 950 lines in 29 files; with its option off, the solver is the released one. The other stages are unchanged: the scene is one CUDA graph, so contact detection, CCD, assembly, the preconditioner build and apply, and the line-search energy evaluations run for the whole scene until its last environment has converged, which takes 1.49\x, 1.70\x, and 1.60\x as many Newton iterations as an environment needs itself on the block press, cloth press, and Franka fold. We repeated the configuration search of Appendix~\ref{app:search} for the modification: its best multi-process packings are those of \cgq ($16{\times}32$, $16{\times}8$, and $16{\times}16$), and its best single scene grows to 256, 128, and 128 environments. Over three interleaved rounds, the modification is 1.05\x (range 1.04--1.05), 1.20\x (1.17--1.20), and 1.09\x (1.08--1.10) faster than \cgq; \system is 8.0\x (7.94--8.01), 8.5\x (8.41\nobreakdash--8.56), and 4.5\x (4.44--4.51) faster than the modification and 12.9\x, 16.8\x, and 6.7\x faster than its best single scene, and the same rounds reproduce the speedups of \cref{tab:benchmarks} within 0.4\%. Per Newton iteration, an environment of the modification performs 15.9, 10.4, and 26.8 PCG iterations (\system: 14.8, 10.2, and 26.8), and each environment-PCG iteration costs 7.4\x, 8.3\x, and 4.6\x as much as in \system, against 5.4\x, 3.9\x, and 2.9\x for \cgq as released (\cref{sec:eval-throughput}).\par
\endgroup

\paragraph{Checks of the modification.} We compare trajectories by the largest vertex deviation over the timing window, sampled every fifth step, on 192 probe environments of each press and 64 of the Franka fold. With one environment per scene, the modification agrees with the released solver about as closely as two runs of the released solver agree with each other: the median deviation is 0.0012\,mm on the block press, 0.28\,mm on the cloth press, and 71\,mm on the chaotic Franka fold, against 0.0012, 0.25, and 79\,mm between two runs. One block-press probe deviates by up to 6.6\,mm (1.8\,mm between two runs), because a step of 23 Newton iterations falls one step later, and agrees to 0.4\,mm afterwards. In a tiled scene, an environment's iteration counts do not depend on its neighbors: replacing a hard neighbor by an easy one leaves the Newton iteration counts of 99.8\% and 95.6\% of the other environments' steps unchanged on the block and cloth presses, as a rerun does (99.7\% and 95.3\%), whereas in \cgq as released 88.9\% and 67.2\% of the scene's steps keep their count. An environment's trajectory still depends on its neighbors: in a tiled scene of the best packing, it deviates from its solo run by a median of 0.16\,mm on the block press, 1.1\,mm on the cloth press, and 87\,mm on the Franka fold. The coupling comes from the MAS preconditioner, whose subdomains mix vertices of different environments, in the released solver too; with a block-Jacobi preconditioner, a tiled block-press scene reproduces the solo runs as closely as a rerun does (post hoc).

\paragraph{Memory traffic.} On identical synchronized inputs, \system reads 50.5\% fewer bytes than \cgq and writes 89.7\% fewer (\cref{fig:ablation}b). At both systems' best configurations, \system moves 72\%, 67\%, and 46\% fewer DRAM bytes per environment step on the block press, cloth press, and Franka fold; part of this gap is \cgq's multi-process configuration, whose 16 processes move 1.4--1.7\x the bytes of a single process. Sharing the nine BSR and MAS index fields saves 544\,MB at $B{=}4{,}096$.

\paragraph{Storage layout.} The environment-outermost build is \system with one switch: every private array is stored with the environment index outermost, and every loop maps adjacent threads to adjacent items of one environment. Each layout-specific choice favors this build. Within an environment it keeps \system's order of items and components, it sums the per-environment dot products of the PCG within a warp, it combines the lanes of a warp that scatter $3{\times}3$ blocks to one address before a single atomic addition, and it traverses the BVH with one thread per query and environment; we did not build a variant with other algorithms, such as a colored or gather-based assembly. At $B^\star$, \system is 1.293\x faster on the block press (range 1.2927--1.2932) and 1.286\x on the cloth press (1.281--1.287), as the median of three interleaved pairs on one H100. Both builds need the same Newton iterations per step and PCG iterations per Newton iteration (2.23 and 14.8 on the block press, 2.38 and 10.2 on the cloth press), so the whole factor is wall time per environment-PCG iteration. With the switch off, the build runs at 1.000 and 0.998 of the default build's throughput, and the median final-state RMS difference between the two builds is 0.29\,\textmu{}m on the block press and 15\,\textmu{}m on the cloth press, the same as between two runs of \system. At $2B^\star$, the best of the batch sizes screened for it on the block press, the outer build remains 1.26\x slower than \system at $B^\star$. Stage profiles of both builds (single runs; the cost of a stage is the time added by executing it twice) locate the difference. Over the timing window, the outer build takes 1.76\x and 1.90\x as long for assembly on the block and cloth presses, 1.73\x and 1.77\x for assembling the preconditioner's cluster matrices, and 2.00\x and 2.11\x for the broad phase, but 0.95\x and 0.83\x for the PCG and 0.77\x and 0.70\x for inverting the cluster matrices. Kernel microbenchmarks on one $B^\star$ state with every environment active, which run the same kernels with transposed private arrays and per-environment index fetches, give a larger effect in the PCG: one PCG iteration is 1.30\x and 1.43\x slower with the environment index outermost at equal DRAM bytes (the achieved bandwidth falls from 2.1 to 1.6\,TB/s), assembly 1.2\x and 3.6\x slower, and the BVH refit 1.06\x; end to end, the PCG stage is not slower.

\paragraph{Contact operator.} Applying the contact factors costs 0.27 and 0.25\,ms per PCG iteration. The cheapest assembled alternative, an unmerged list of dense per-stencil $12{\times}12$ blocks as GIPC-lineage solvers keep it, applies in 0.68 and 0.44\,ms, reads 4.7--7.6\x the DRAM bytes, and must be rebuilt every Newton iteration; per Newton iteration the factors are 2.8\x and 2.0\x cheaper. Merging the blocks into sorted per-environment patterns makes an application faster (0.11\,ms) but its rebuild slower (6.9--9.4\,ms), so it pays off only beyond about 50 PCG iterations per Newton iteration; the presses average 10--15 and the Franka fold 27.

\paragraph{Stage shares.} After all optimizations, PCG accounts for 44--47\% of \system's time on the block press, preconditioner construction for 21\%, assembly for 14\%, collision detection for 13--15\%, and CCD with line search for about 4\%.

\section{Regime details}
\label{app:regime}
This appendix supports \cref{sec:eval-hetero,sec:eval-generality}.

\paragraph{Resolution sweep.} With synchronized contact, per-environment decisions on the cloth press still save 1.45--2.2\x of the PCG iterations up to 2,883 DOF (1.1\x at 12,675 DOF), and iterations are 3.6--5.4\x cheaper. At 49,152 DOF the cloth's edges (2.4\,mm) are about as long as $\hat d$ (2\,mm), so even lying flat it carries about 129,000 active self-contact pairs per environment; both systems take the same iterations, and with randomized contact \system's environments desynchronize inside the batched PCG loop, which then needs 1.6\x as many trips for the same work, while a trip costs the same however few environments still iterate---the remaining synchronization of \cref{sec:scheduling}; \system is then 1.03\x as fast as \cgq with randomized contact, against 1.66\x with synchronized contact.

\paragraph{Cloth drop.} On the cloth drop and the $65^2$ press, per-environment decisions save little (1.15\x and 1.09\x fewer PCG iterations per environment step). After landing, the dropped cloth lies in folds with about 1,300--1,500 active self-contact pairs per environment. There, \system's 1.73\x (refill) and 1.77\x (lockstep) over \cgq (\cref{sec:eval-hetero}) come from two independent sets of paired runs, and \cgq's single scene fails beyond $B{=}64$ because its edge--edge collision traversal exhausts a fixed-size stack.

\paragraph{Periodic-contact benchmark.} On the periodic-contact benchmark with randomized contact phases, matching contact-buffer capacity between the modes changes the cloth ratio by less than 1\% (2.27\x).

\par\begingroup
\ifdim\pagegoal<\dimexpr\pagetotal+6\baselineskip\relax
  \newpage
\fi
\endgroup
\section{Application details}
\label{app:applications}
This appendix supports \cref{sec:applications}.

\paragraph{Start-up.} With both backends' start-up included, the pick-cube estimate reaches $\pm1\%$ 2.9\x sooner with a warm cache and 1.3\x from a cold start (3.7\x without), and the three pick, peg and cable lists finish 7.5--9.7\x and 4.4--7.5\x sooner (6.4--12.7\x without).

\paragraph{Capped steps.} In policy evaluation, steps that reach the Newton cap are 0.0009\% of \system's steps on the pick cube (\cgq: at most 0.006\%), one in 31 million on the peg insertion (\cgq: at most 0.0003\%), none on cable routing for either backend, and 1.3\% on the LeHome garment fold (\cgq: at most 1.5\%). \cgq's fractions of Newton-capped steps in policy evaluation are upper bounds read from the logs of the timed runs; \system's come from reruns of the same configurations that read out its per-environment budget counter. On the $\pi_0$ towel, none of \system's 384,000 steps is Newton-capped, and \cgq's count was not recorded. In the optimization searches of \cref{tab:applications}, Newton-capped steps are 37 of 761 million environment steps with \system (fewer than one in ten million on each task) and 60 of 58 million scene steps with \cgq (none on the rope gathering, 0.4 and 2.6 per million on the slingshot and the ring wiring); because \cgq tests convergence once per scene, each of its capped steps discards all 16 candidates of the scene. Contact buffers overflowed in a few steps of \system's runs of \cref{tab:applications} (for example, in 36 of the 97 million environment steps of its five cloth-fold PPO runs); each was rolled back and replayed with grown buffers, and in these runs no step on either backend was kept after an overflow or ended an episode as invalid.

\paragraph{Policy evaluation.} In \cref{tab:applications}, \system uses the faster of refill and lockstep: refill for the pick cube, the peg, and the garment fold, lockstep for the cable; the towel ran in lockstep only. The fastest healthy \cgq packings (processes $\times$ environments per process) are $16{\times}8$ for the pick cube, $16{\times}64$ for the peg, $4{\times}128$ for the cable, and $16{\times}1$ for the garment fold. Quality agrees within 1.1 percentage points on the three rows without path tracing; the $\pi_0$ towel reports execution cost only. Garment-fold quality in the table is the success rate over the timed list of 384 episodes per backend; on the official 60-episode list, \cgq loses 15 of 120 episodes to a native error. Measured quality in \system differs from the native simulator's where the contact models differ: the SARNN policy, trained in MuJoCo~\citep{todorov2012mujoco}, succeeds in 241 of 256 episodes in \system (94.1\%) and in 210 of 256 natively (82.0\%), because \system's IPC rod slides between the poles more easily than MuJoCo's hinge chain; and an ACT policy~\citep{zhao2023aloha} for LeHome's long-sleeve folding task~\citep{li2026lehome} succeeds in 30 and 22 of 60 episodes in two \system runs and in 31 of 60 in the official evaluator from the same initial states; this comparison establishes comparable success rates on a task whose outcomes are not repeatable episode by episode, not per-episode reproduction. The official LeHome evaluator ran on Isaac Sim 6.0 because the policy's Isaac Sim 5.1 does not run on our GPU driver. Its per-episode outcomes agree with \system's in 25 of 60 episodes, no more than the 30 expected by chance (Cohen's $\kappa{=}{-}0.17$), and two \system runs agree in 40 of 60 ($\kappa{=}0.33$). The gain over \cgq is smallest for the peg insertion, on which \cgq's tiled scene keeps scaling to 256 environments per process. With a lighter rasterizer, the end-to-end gain of the visual $\pi_0$ policy is 2.0\x.

\paragraph{Derivative-free trajectory optimization.} In the \emph{rope gathering}, a rope driven at both ends must herd five blocks---four affine bodies and one soft finite-element box---through a 12\,cm gap in a wall onto a 6.2\,cm disc behind it and leave all five at rest there (43 parameters, 7\,s). In the \emph{narrow-gate slingshot}, an elastic rod clamped at both ends launches a ball through two windows with 3\,mm of clearance into a cup 37\,cm away (29 parameters, 2.5\,s). In the \emph{ring wiring}, a 0.9\,m cable must thread three rings tilted by 40--45$^\circ$ with 4.2\,cm apertures in order and in the required direction (56 parameters, 6.8\,s). Rods use discrete elastic rods~\citep{bergou2008der} coupled to IPC contact; the gathering and slingshot use block-Jacobi-preconditioned CG, the wiring MAS. For the rope gathering, passing every gate means that all five blocks pass through the gap and come to rest on the disc; there \cgq's default buffer growth stalls the GPU, so we preset its buffers.

On the slingshot, \cgq's searches need more generations on average (48.0 against 29.0; two took 121 and 128, against at most 64 with \system), but single searches vary so much (1 to 64 generations with \system) that the difference is not significant (ratio 1.66, 95\% interval 0.82--3.04). On the rope gathering, \cgq's searches need fewer generations (9.8 against 15.0; ratio 0.65, interval 0.59--0.74; \mbox{Mann--Whitney} $p{=}0.015$), so the time ratio stays below the throughput ratio. The two simulators do not serve a search identically.

On the same 8,192 candidates per task, their guide losses agree in rank (Spearman correlations 0.92, 0.83, and 0.97, in the task order above) but differ by more than either backend varies between two of its own runs: on the rope gathering, the median relative difference is 7.1\%, against 1.0\% between two \system runs and 0.2\% between two \cgq runs. There the difference also has a direction: \cgq's losses are lower, by a median of 3.3\% on the 6,135 candidates with losses between 0.01 and 0.05, and 67--74 candidates fall below the target loss in each of its six runs, against 39--52 with \system. This shift comes from convergence, not from the model. \cgq applies the stopping test to all 16 environments of a scene, so most environments iterate past the tolerance at which \system stops each of them (7.5 Newton iterations per step, against 4.4 when \cgq simulates one environment per scene). With a ten times tighter tolerance, \system reproduces \cgq's losses (median difference below 0.1\%; 76 candidates below the target loss), and \cgq with one environment per scene reproduces \system's on a sample of 384 candidates. At that tolerance, \system's rate on the rope candidates falls to 0.60 times its default rate (0.59--0.62 in three paired runs), which puts its throughput ratio to \cgq at 6.9\x instead of 11.6\x. Five searches at that tolerance, with the same seeds, need 12.2 generations on average instead of 15.0 and reach the target 6.8\x sooner than \cgq's (95\% interval 6.3--7.3\x). \cgq's searches still need fewer generations (9.8; ratio 0.80, interval 0.71--0.91; Mann--Whitney $p{=}0.043$), so matched convergence accounts for about half of the rope's generation difference; we have not identified the rest. The slingshot and ring-wiring losses show no such shift (below 0.1\%).

The backends also disagree on slingshot candidates near the target: of the 20,480 first-generation candidates of the ten slingshot seeds, which are the same on both backends, \system simulates two as meeting the target and \cgq none.

The nearly equal \cgq mean times on the rope gathering and slingshot (239.79 and 239.78\,min) are a coincidence. \cgq also loses time to native CUDA errors and stalls (62 and 27 in its 25 searches; \system had none): we retry a failed wave once from the last completed generation, record its candidates as failed if it fails again, and charge all of this time, which is 1\%, 20\%, and 19\% of \cgq's search time on the three tasks; without it, the time ratios would be 7.4\x, 4.5\x, and 3.0\x. A found candidate is a search result, not a certificate: simulated again in a batch of other candidates, the target candidate of a development search meets the target again on the ring wiring with both backends, on the slingshot with \system only, and on the rope gathering with neither.

\paragraph{Reinforcement learning.} The pick-cube collector comparison runs both systems at 1,024 environments in lockstep with a stochastic policy, whereas evaluation (\cref{tab:applications}) uses each system's best configuration, where \system gains 3.3\x from batch size and refill and \cgq 1.4\x from packing and a deterministic actor. Over the updates before the target is reached, with the training runs' own policies, the collector alone is 8.1\x faster, and the identical policy-update cost on both backends dilutes this to 7.9\x; over the complete training runs, the rate of \cref{sec:applications} is 7.8\x. In the timed cloth-fold training, with the same policy, \cgq's fastest configuration simulates 3.6\x more slowly, and its remaining per-sample work is 7.7\x larger, because a 32-environment scene needs sixteen times as many policy and environment calls per sample as \system's 512 environments. In these runs, 0.05\% of \system's PCG solves reach their cap. Cross-evaluation on the same episodes (\system{}\,/\,\cgq{}) gives pick-cube success of 95.9\%\,/\,96.1\% for \system-trained policies and 95.7\%\,/\,95.7\% for \cgq-trained ones, and cloth-fold success of 92.7\%\,/\,86.9\% and 91.9\%\,/\,91.1\%, respectively. Each time to target measures the same success on the training simulator's own dynamics.

\clearpage

\ifbqdraft\input{sections/99_register}\fi
\end{document}